\documentclass[letterpaper]{article} 
\usepackage{aaai25}  
\usepackage{times}  
\usepackage{helvet}  
\usepackage{courier}  
\usepackage[hyphens]{url}  
\usepackage{graphicx} 
\usepackage{natbib}  
\usepackage{caption} 
\usepackage{algorithm}
\usepackage{algorithmic}

\usepackage{newfloat}
\usepackage{listings}
\DeclareCaptionStyle{ruled}{labelfont=normalfont,labelsep=colon,strut=off} 
\floatstyle{ruled}
\newfloat{listing}{tb}{lst}{}
\floatname{listing}{Listing}
\usepackage{marvosym}

\title{ZipGait: Bridging Skeleton and Silhouette with Diffusion Model \\ for Advancing Gait Recognition
}
\author{
    Fanxu Min,
    Qing Cai,
    Shaoxiang Guo,
    Yang Yu,
    Fan Hao,
    Junyu Dong
}
\affiliations{
    School of Information Science and Engineering, Ocean University of China,
}

\makeatletter
\newcommand{\maketitleB}{
  \begin{center}
    {\bfseries\LARGE \@title \par} 
    \vskip 2em 
  \end{center}
}
\makeatother

\usepackage{bibentry}
\usepackage{diagbox}  
\usepackage{multirow} 
\usepackage{booktabs} 
\usepackage{enumitem}
\usepackage{subcaption}
\setlist[itemize]{leftmargin=*}
\def\blue#1{\textcolor{blue}{#1}}
\usepackage{xcolor}
\usepackage{amssymb}
\usepackage{colortbl} 
\usepackage{tabularx} 
\usepackage{amsmath}

\usepackage{pifont}

\begin{document}

\maketitle

\begin{abstract}
Current gait recognition research predominantly focuses on extracting appearance features effectively, but the performance is severely compromised by the vulnerability of silhouettes under unconstrained scenes.
Consequently, numerous studies have explored how to harness information from various models, particularly by sufficiently utilizing the intrinsic information of skeleton sequences.
While these model-based methods have achieved significant performance, there is still a huge gap compared to appearance-based methods, which implies the potential value of bridging silhouettes and skeletons.
In this work, we make the first attempt to reconstruct dense body shapes from discrete skeleton distributions via the diffusion model, demonstrating a new approach that connects cross-modal features rather than focusing solely on intrinsic features to improve model-based methods.
To realize this idea, we propose a novel gait diffusion model named DiffGait, which has been designed with four specific adaptations suitable for gait recognition.
Furthermore, to effectively utilize the reconstructed silhouettes and skeletons, we introduce Perception Gait Integration (PGI) to integrate different gait features through a two-stage process.
Incorporating those modifications leads to an efficient model-based gait recognition framework called \textbf{ZipGait}.
Through extensive experiments on four public benchmarks, ZipGait demonstrates superior performance, outperforming the state-of-the-art methods by a large margin under both cross-domain and intra-domain settings, while achieving significant plug-and-play performance improvements.
\end{abstract}
%
\section{1 Introduction}
Gait recognition, as a biometric technology, facilitates remote identification in uncontrolled settings without necessitating subject participation, discerning individuals based on their gait patterns \cite{1-wu2016comprehensive, 2-wang2003silhouette}.
Current research focuses on utilizing shape information of silhouettes for recognition, known as appearance-based methods \cite{39-dou2024clash, 43-dou2023gaitgci, 44-wang2023dygait, 51-lin2021gaitgl, 45-zheng2022gait}.
However, these methods are vulnerable to complex backgrounds, severe occlusion, arbitrary viewpoints, and diverse clothing changes.
This gives rise to a surging interest in model-based methods that extract gait features through various models, including 2D/3D Skeleton, SMPL model, and Point cloud \cite{10-loper2023smpl, 56-yam2004automated, 41-shen2023lidargait}.

\begin{figure}[!t]
    \setlength{\abovecaptionskip}{2pt}
    \centering
    \includegraphics[width=0.95\linewidth]{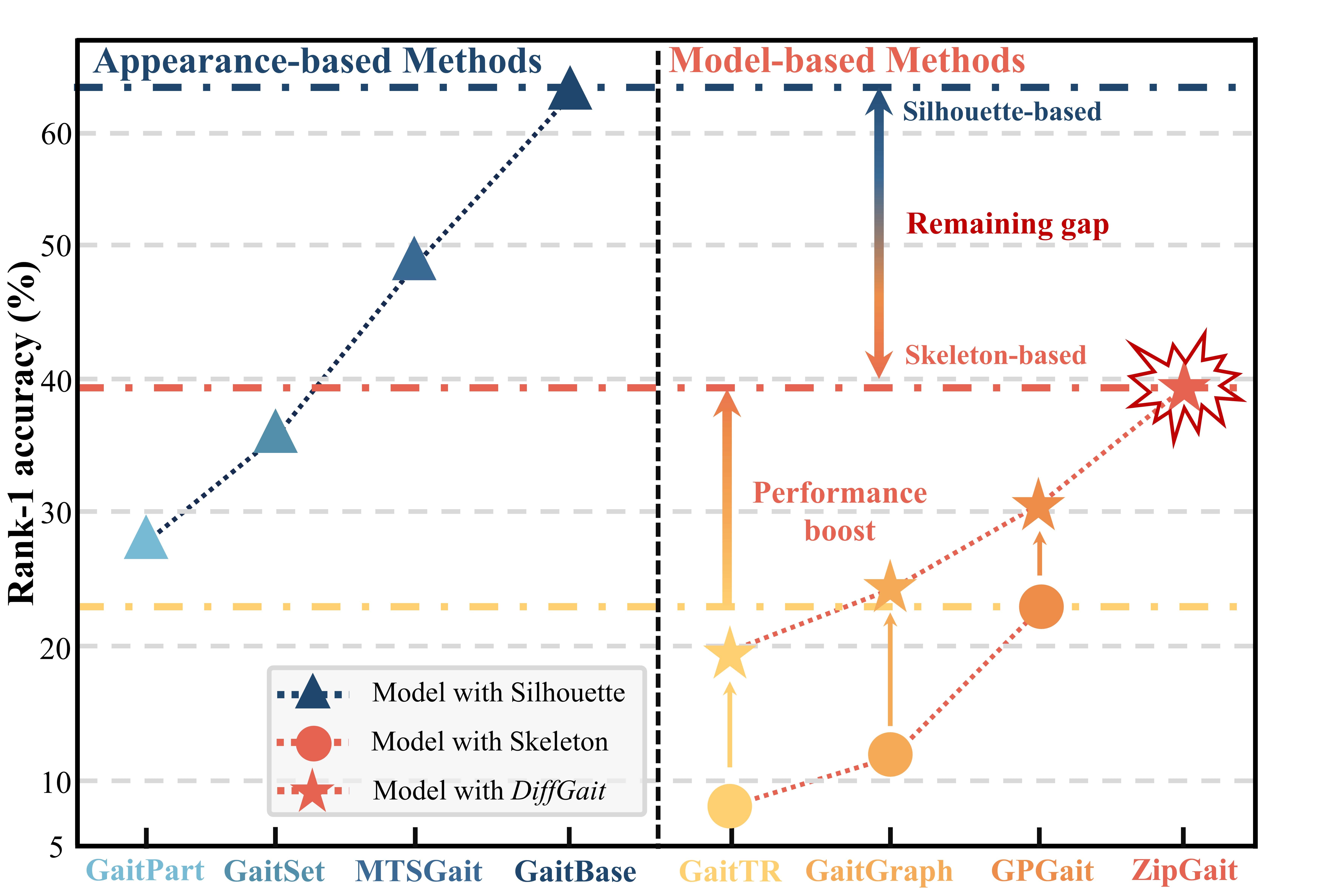}
    \caption{
    \small Comparison with alternative methods on the Gait3D test set elucidates the performance disparities between two types of methods in real-world scenarios. 
    DiffGait enhances model-based methods through a plug-and-play approach, reducing the performance gap.
    Meanwhile, ZipGait achieves the best-performing compared to methods using skeletons.
    }
    \label{performance}
    \vspace{-0.5 cm}
\end{figure}

Among them, 2D skeleton is particularly attractive since it achieves higher pose estimation accuracy due to low spatial complexity which demonstrates simplicity and effectiveness \cite{9-liao2020model, 52-li2020end, 6-teepe2021gaitgraph}.
Most existing model-based works, using the skeleton as input, focus on extracting intrinsic information from skeleton sequences to provide sufficient features, broadly categorized as follows: 
utilization of skeleton structural features (i.e. position, angle, length, local-global \cite{9-liao2020model});
computation of physical information from skeleton sequences (i.e. motion velocity, gait periodicity \cite{59-frank2010activity, 62-liu2022symmetry}); 
generation of complementary skeleton information (i.e. multi-view, frame-level correlations \cite{58-gao2022gait, 57-wang2022frame}).

\begin{figure}[!t]
    \setlength{\abovecaptionskip}{2pt}
    \setlength{\belowcaptionskip}{2pt}
    \centering
    \includegraphics[width=0.95\linewidth]{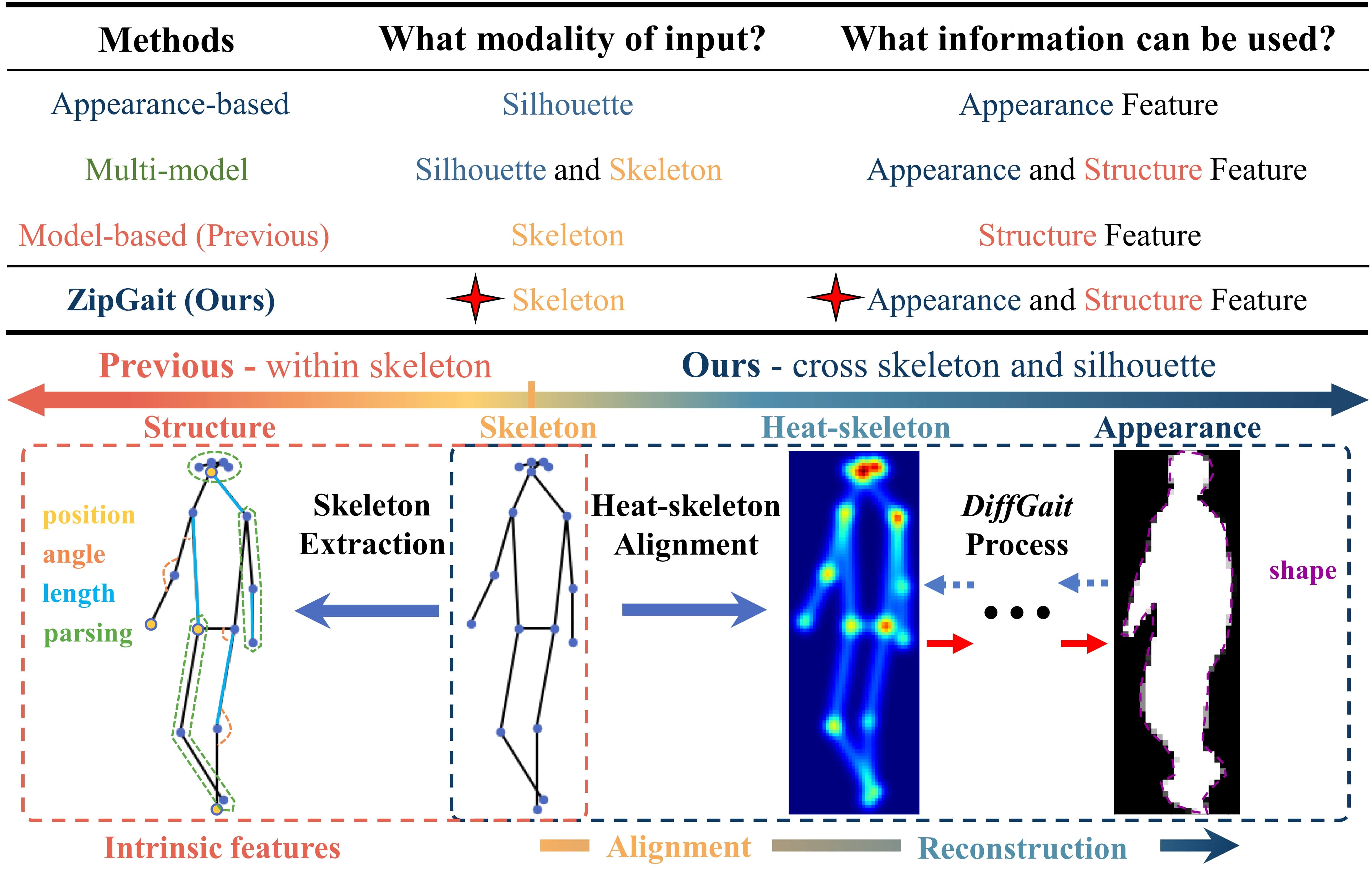}
    \caption{\small Differences between our approach and related gait recognition methods.
    Comparison of our cross-modal insight with previous works that primarily focused on skeleton extraction.
    }
    \vspace{-5 mm}
    \label{diffusion_overview}
\end{figure}

Although existing model-based methods have shown progressive improvements in performance, they continue to lag behind appearance-based approaches, suggesting that shape features can offer more gait imformation, as indicated in Fig~\ref{performance}.
This observation prompts us to consider whether merely utilizing intrinsic information of skeletons could bridge this huge gap between these two types of methods.
We propose a novel concept: establishing a natural correlation between silhouettes and skeletons to reconstruct dense body shapes from sparse skeleton structures, inspired by the progress of diffusion models \cite{15-ho2020denoising, 16-song2020denoising}.
As shown in Fig.~\ref{diffusion_overview}, instead of solely leveraging information from intra-modality gait features, we shift towards connecting cross-modality gait features for a more effective gait representation, thereby enhancing the performance of model-based methods.

To realize the aforementioned idea, we introduce a novel \underline{Gait} \underline{Diff}usion Model named \textbf{DiffGait}, which denoises the skeleton features mixed with standard Gaussian noise using a diffusion model to retrieve matching silhouettes.
Intuitively, several limitations hinder the diffusion model's direct application in gait recognition, which DiffGait addresses with four aspects:
\textbf{1)} Given that silhouettes and skeletons represent two unalignable modalities that inherently resist direct transformation via the diffusion process, we employ Heat-skeleton Alignment to transform 2D skeleton joints into a unified mesh grid, bridging skeletons and silhouettes across spatial-temporal dimensions;
\textbf{2)} To address the cumbersome algorithms of the standard diffusion process for gait recognition, the DiffGait Forward Process has been developed to convert the diffusion object from the silhouette to the Hybrid Gait Volume (HGV), streamlining the condition and noise encoding stages;
\textbf{3)} Recognizing that existing diffusion models neglect the intrinsic correspondence between modalities during the denoising process, we implement the DiffGait Reverse Process to produce multi-level silhouettes, which improves the distinction in denoising outcomes;
\textbf{4)} Mainstream diffusion models, predominantly based on U-Net \cite{64-ronneberger2015u}, are unsuitable for gait recognition due to their large scale. 
In response, our DiffGait Architecture adopts a decoder-only configuration to achieve a surprisingly efficient model with just 1.9M parameters and a rapid inference speed of 3628 FPS.

Furthermore, to efficiently leverage appearance and structure features, we propose a gait feature fusion module termed \textbf{Perceptual Gait Integration (PGI)}, which operates through a two-stage process that refines silhouettes and extracts hybrid gait features respectively.
By incorporating these two designs into our baseline, we create a simple-but-effective model-based gait recognition architecture, named \textbf{ZipGait}, which seamlessly bridges skeleton and silhouette like a zipper.
Fig.~\ref{diffusion_overview} summarizes the distinctions between our proposed method and existing gait recognition approaches, which are further discussed in detail in the Appendix. C.
Extensive experiments on four dominant datasets demonstrate that our method outperforms state-of-the-art approaches in both cross-domain and intra-domain settings while achieving significant plug-and-play performance improvements.

\begin{figure*}[ht]
    \setlength{\abovecaptionskip}{2pt}
    \setlength{\belowcaptionskip}{2pt}
    \begin{center}
        \includegraphics[width=0.95\linewidth]{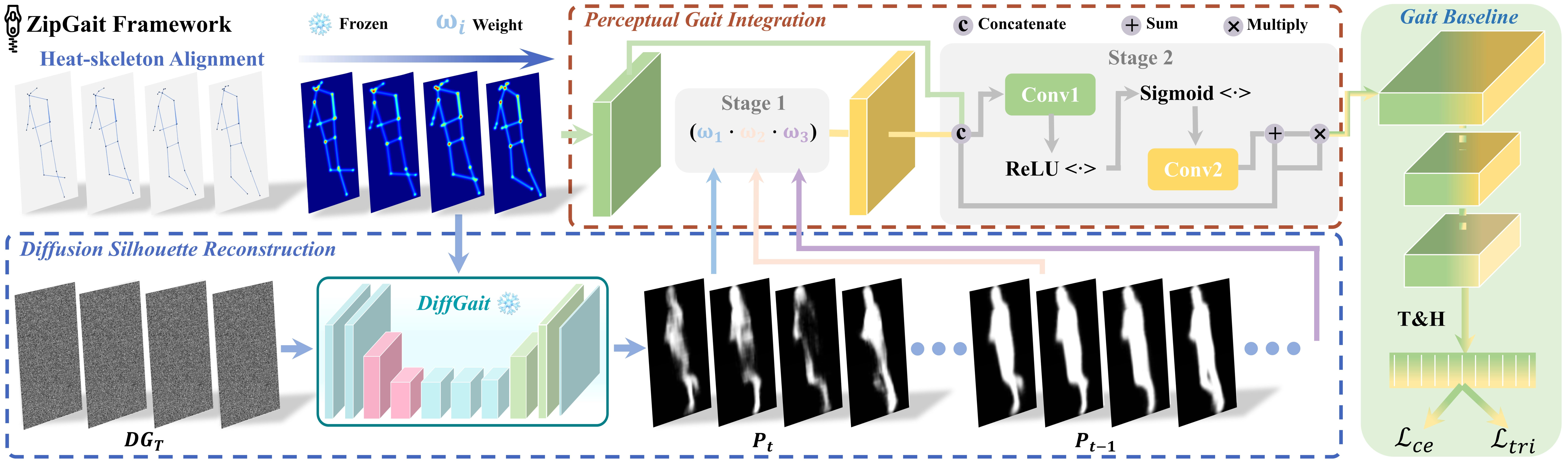}
    \end{center}
    \caption{\small Overall pipeline of the proposed ZipGait framework.
    It consists of two fundamental improvements: DiffGait and Perceptual Gait Integration.
    The entire denoising process of DiffGait is summarized as Diffusion Silhouette Reconstruction.
    \textbf{T\&H} represents the horizontal mapping \cite{29-fu2019horizontal} and temporal aggregation.}
    \label{ZipGait}
    \vspace{-0.7cm}
\end{figure*}  

\section{2 Related Work}
\noindent \textbf{Model-based Gait Recognition.} 
These methods utilize the underlying structure of the human body as input.
In particular, skeleton representation is enhanced by extracting intrinsic information from skeletal sequences.
GaitGraph2 \cite{7-teepe2022towards} pre-computes skeletons to obtain joint positions, motion velocities, and length of bones as gait features.
GPGait \cite{8-fu2023gpgait} transforms the arbitrary human pose into a unified representation and allows efficient partitions of the human graph.
\cite{13-fan2024skeletongait, 63-Min2024GaitMAPM} represents the coordinates of human joints as a heatmap to provide explicit structural features.
\cite{31-li2022strong, 30-choi2019skeleton} utilize gait periodicity priors and frame-level discriminative power, respectively.
\cite{32-huang2023condition} dynamically adapts to the specific attributes of each skeleton sequence and the corresponding view angle.
\cite{33-pan2023toward} generate multi-view pose sequences for each single-view sample to reduce the cross-view variance.
Existing studies have significantly improved the extraction of efficient gait features from skeletons.
However, none has explored the connection between structure and appearance to improve the performance of model-based methods.

\noindent \textbf{Appearance-based Gait Recognition.} 
These methods attempt to learn gait features directly from silhouette sequences, which has a big performance gap compared with model-based methods. 
GaitSet \cite{3-chao2019gaitset} innovatively regarded the gait sequence as a set and compressed frame-level spatial features. 
GaitPart \cite{5-fan2020gaitpart} carefully explored the local details of the input silhouette and modeled the temporal dependencies. 
GaitBase \cite{4-fan2023opengait} is an efficient baseline model that demonstrates applicability across various frameworks and gait modalities.
MTSGait \cite{45-zheng2022gait} learns spatial features and multi-scale temporal features simultaneously.
This demonstrates the feasibility of bridging the gap between skeletons and silhouettes to improve the model-based methods.

Additionally, recent works aim to effectively integrate gait modalities to enhance recognition performance \cite{18-peng2024learning, 19-cui2023multi, 20-dong2024hybridgait, 21-zheng2022gait}, and some research is devoted to obtaining richer gait representations \cite{14-guo2023gaitcontour, 53-pinyoanuntapong2023gaitmixer, 55-han2005individual, 40-wang2023gaitparsing, 12-zheng2023parsing, 60-zou2024cross, 61-wang2024qagait}. 
Our proposed ZipGait diverges from these multimodal methods that use both silhouettes and skeletons as inputs, we only use skeletons.
For a detailed comparison, see the Appendix. C.

\noindent \textbf{Diffusion Model.}
Diffusion models, also widely known as DDPM, are a series of models generated through Markov chains trained via variational inference \cite{15-ho2020denoising}. 
\cite{16-song2020denoising} replaced the Markovian forward process used in earlier studies with a non-Markovian process, introducing the well-known DDIM. 
LDM \cite{17-rombach2022high} has achieved image generation guided by textual inputs, advancing research in cross-modal generation using diffusion models. 
Diffusion-based methods have achieved impressive results in generation and segmentation tasks \cite{48-chen2023diffusiondet, 47-feng2023diffpose, 46-holmquist2023diffpose, 23-gong2023diffpose}.
Inspired by these studies, we have introduced diffusion models to generate dense body shapes from sparse skeleton distributions, thereby establishing a direct relational model between gait modalities for the first time.

\section{3 Method}
\subsection{3.1 Problem Formulation}
Denote a human skeleton sequence with total $N$ frames as $\mathcal{V}_s = \{\mathcal{S}_t\}_{t=1}^N$, where $\mathcal{S}_t \in \mathbb{R}^{ kp \times 3}$ refers to the $t$-th frame, previous methods identify individuals based on skeleton structure features $\mathcal{F}_s$ within the video sequence $\mathcal{V}_s$.
Our work is the first attempt to establish the natural correlation between appearance and structure features to eliminate discrepancies between silhouette and skeleton, which can easily reconstruct silhouette appearance features $\mathcal{F}_a$ from $\mathcal{F}_s$ and achieve excellent performance.

\subsection{3.2 ZipGait Workflow}
Inspired by \cite{63-Min2024GaitMAPM, 25-duan2022revisiting, 13-fan2024skeletongait}, our work employs heatmaps to represent skeletons, termed 'Heat-skeleton'. 
Leveraging the OpenGait \cite{4-fan2023opengait} framework, we have optimized its structure for enhanced fine-grained gait fusion, as shown in Fig.~\ref{ZipGait}.

Given a skeleton frame $\mathcal{S}_t$, it is first transformed into Heat-skeletons $\mathcal{I}_t$. 
Subsequently, $\mathcal{I}_t$ is input into \textit{DiffGait} to reconstruct the corresponding silhouettes.
DiffGait Reverse Process could generate multi-level silhouettes under different sampling steps. 
This process undergoes M rounds of sampling to predict $\{ P_i \}_{i=1}^{M}$, which we regard as Diffusion Silhouette Reconstruction.

$\mathcal{I}_t$, $\{ P_i \}_{i=1}^{M}$ serve as inputs to Perceptual Gait Integration to obtain a comprehensive gait representation. 
In stage one, $\{ P_i \}_{i=1}^{M}$ are then refined through dynamic weight allocation to yield the final composite silhouette $\mathcal{P}_t$.
In stage two, after convolutional initialization, $\mathcal{P}_t$ and $\mathcal{I}_t$ are blended through gait fusion layer to obtain the hybrid gait feature $\mathcal{H}_t$.

Finally, $\mathcal{H}_t$ is processed through GaitBase \cite{36-he2016deep, 4-fan2023opengait} to generate the predicted identity as a one-hot vector.
During the training phase, the model's learning process is supervised by calculating both the triplet loss \cite{38-hermans2017defense} and the cross-entropy loss, which can be formulated as:
\begin{eqnarray}
    {\cal L}_{\mathrm{ce}}&=&-{\frac{1}{N}}\sum_{i=1}^{N}log{\frac{e^{W_{y_{i}}^{T}F_{i}+b_{y_{i}}}}{\sum_{j=1}^{n}e^{W_{j}^{T}F_{i}+b_{j}}}}, 
\end{eqnarray}
\begin{eqnarray}
    {\cal L}_{tri}&=&\varphi \left [ \mathcal{D}\left ( F_{i},F_{k}  \right ) - \mathcal{D}\left ( F_{i},F_{j} \right ) +m \right ],
\end{eqnarray}
$F_{i}$, $F_{k}$ are the features of sample i and sample k, $\varphi$ is equal to $max\left ( \alpha, 0 \right )$, $\mathcal{D}$ represents the Euclidean distance, m denotes the margin for the triplet loss.

\begin{figure*}[ht]
    \setlength{\abovecaptionskip}{2pt}
    \setlength{\belowcaptionskip}{2pt}
    \begin{center}
        \includegraphics[width=0.95\linewidth]{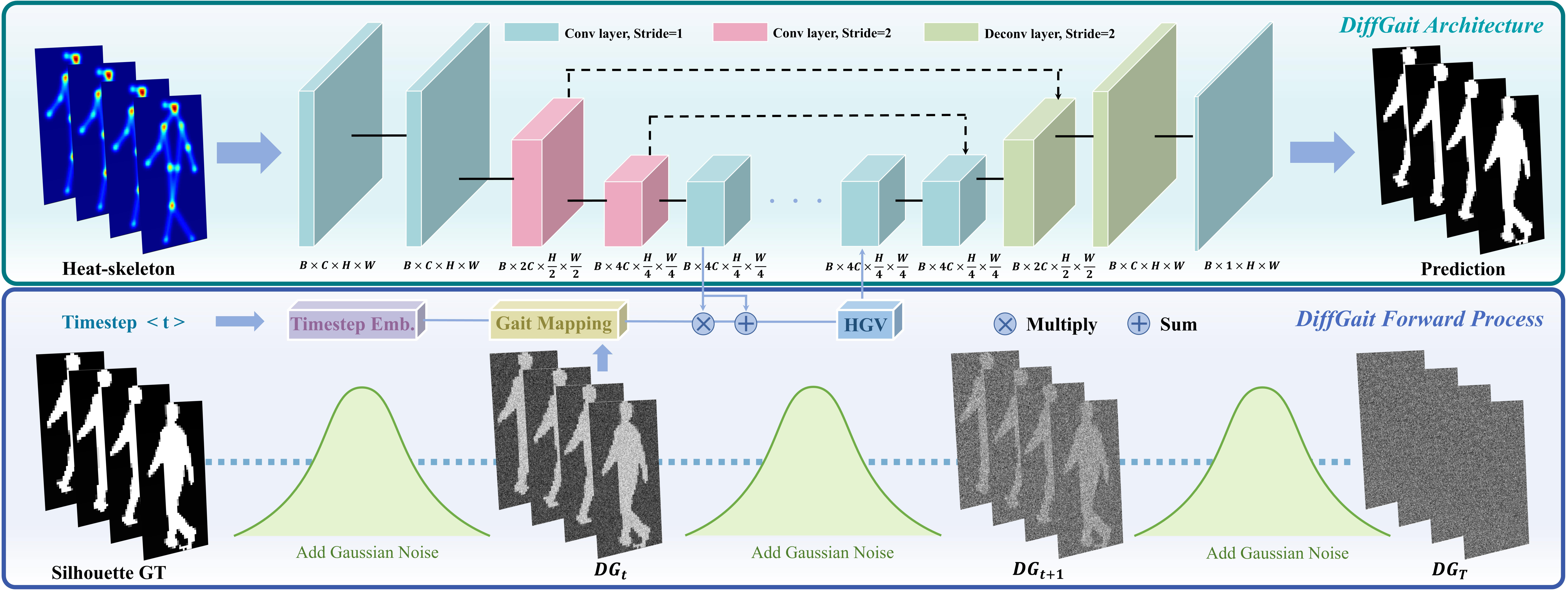}
    \end{center}
    \caption{\small The upper half of the diagram delineates the specific network architecture and modules of DiffGait, while the lower half illustrates the forward process of DiffGait. 
    The entire figure provides a detailed explanation of the feature flow process within DiffGait.}
    \label{DiffGait}
    \vspace{-0.35cm}
\end{figure*}  

\subsection{3.3 DiffGait}
We designed a pioneering diffusion model, termed DiffGait, aimed at reconstructing dense body shapes from sparse skeleton structures. 
However, directly applying diffusion models for this task inevitably encounters limitations due to modality inconsistency, non-specific diffusion process and model scale.
To handle these issues, we initially employ Heat-skeleton Alignment to unify the modalities; subsequently, we design the DiffGait Process consisting of two opposite processes: the DiffGait Forward Process and the DiffGait Reverse Process; ultimately, we develop a highly streamlined DiffGait Architecture, as shown in Fig.~\ref{DiffGait}.

\noindent \textbf{Heat-skeleton Alignment.}
This improvement focuses on two aspects: \textbf{1)} Heat-skeletons explicitly provide spatial structure features; \textbf{2)} Heat-skeletons and silhouettes share modality consistency.

According to the method proposed by \cite{25-duan2022revisiting}, the joint-based heatmap $\mathcal{J}$ centered on each skeleton point is created using the coordinate triplets $\left( x_{k}, y_{k}, c_{k} \right)$:
\begin{eqnarray}
    \label{1}
    \mathcal{J}_{k i j}&=&e^{-\frac{(i-x_{k})^{2}+(j-y_{k})^{2}}{2*\sigma^{2}}}\ast c_{k},
\end{eqnarray}
$\sigma$ regulates the variance of the Gaussian maps, while $\left( x_{k}, y_{k} \right)$ represents the spatial location of the $k$-th joint, and $c_{k}$ represents the corresponding confidence score. We can also create the limb-based heatmap $\mathcal{L}$:
\begin{eqnarray}
    \label{2}
    {\mathcal{L}}_{k i j}\,&=&\,e^{-\frac{D((i,j),s e g[a_{k},b_{k}])^{2}}{2\ast\sigma^{2}}}\ast\mathrm{min}\big({\mathcal{C}}_{a_{k}}\,,\,{\mathcal{C}}_{b_{k}}\big).
\end{eqnarray}
The limb indexed as k connects two joints, $a_{k}$ and $b_{k}$. The function ${\cal D}$ calculates the distance from the point $\left( i, j \right)$ to the segment $\left[\left(x_{a_{k}},y_{a_{k}}\right),\left(x_{b_{k}},y_{b_{k}}\right)\right]$.

\noindent \textbf{DiffGait Forward Process.}
To simplify the diffusion process, we convert the diffusion object from silhouettes to Hybrid Gait Volume (HGV).
Given a Heat-skeleton sequence with total $N$ frames as $\mathcal{V}_h = \{\mathcal{I}_t\}_{t=1}^N$, where $\mathcal{I}_t \in \mathbb{R}^{h \times w \times 2}$ refers to the $t$-th frame.

To obtain $DG_t$, we designate the silhouettes aligned with Heat-skeletons as the diffusion target and define them as the initial state $DG_0$ of the diffusion process. 
We introduce Gaussian noise at various timesteps to facilitate the forward diffusion of the silhouettes,
\begin{equation}
    DG_t=\sqrt{\overline{\alpha}_t}DG_0+\sqrt{1-\overline{\alpha}_t}\epsilon.
\end{equation}

Subsequently, $G_{ske} \in \mathbb{R}^{C \times H/4 \times W/4}$ is extracted by $\mathcal{E}$. 
For dimension matching, we apply \textit{Gait Mapping} to reduce $DG_t$ by a factor of four via a two-layer convolution and increase feature channels to $C$. 
This approach retains more details compared to the direct bilinear interpolation used in DiffuVolume \cite{24-zheng2023diffuvolume}.

Furthermore, timestep embedding of dimension $C$ corresponding to $t$ is generated by \textit{Timestep Embedding} and is added to $DG_t$, then we match the features of $DG_t$ and $G_{ske}$ by element-wise multiplication and introduce skip connections to ensure a smooth transition of the features,
\begin{equation}
    HGV_t=G_{ske}\odot(GM(DG_{t})+TE(t)) + G_{ske},
    \label{hgv}
\end{equation}
where $\odot$ is the element-wise multiplication, $HGV_t$ means the Hybrid Gait Volume, $GM$ means the Gait Mapping, $t$ is the selected timestep and $TE$ means Timestep Embedding. 

\begin{figure}[h]
    \setlength{\abovecaptionskip}{2pt}
    \setlength{\belowcaptionskip}{2pt}
    \centering
    \includegraphics[width=0.8\linewidth]{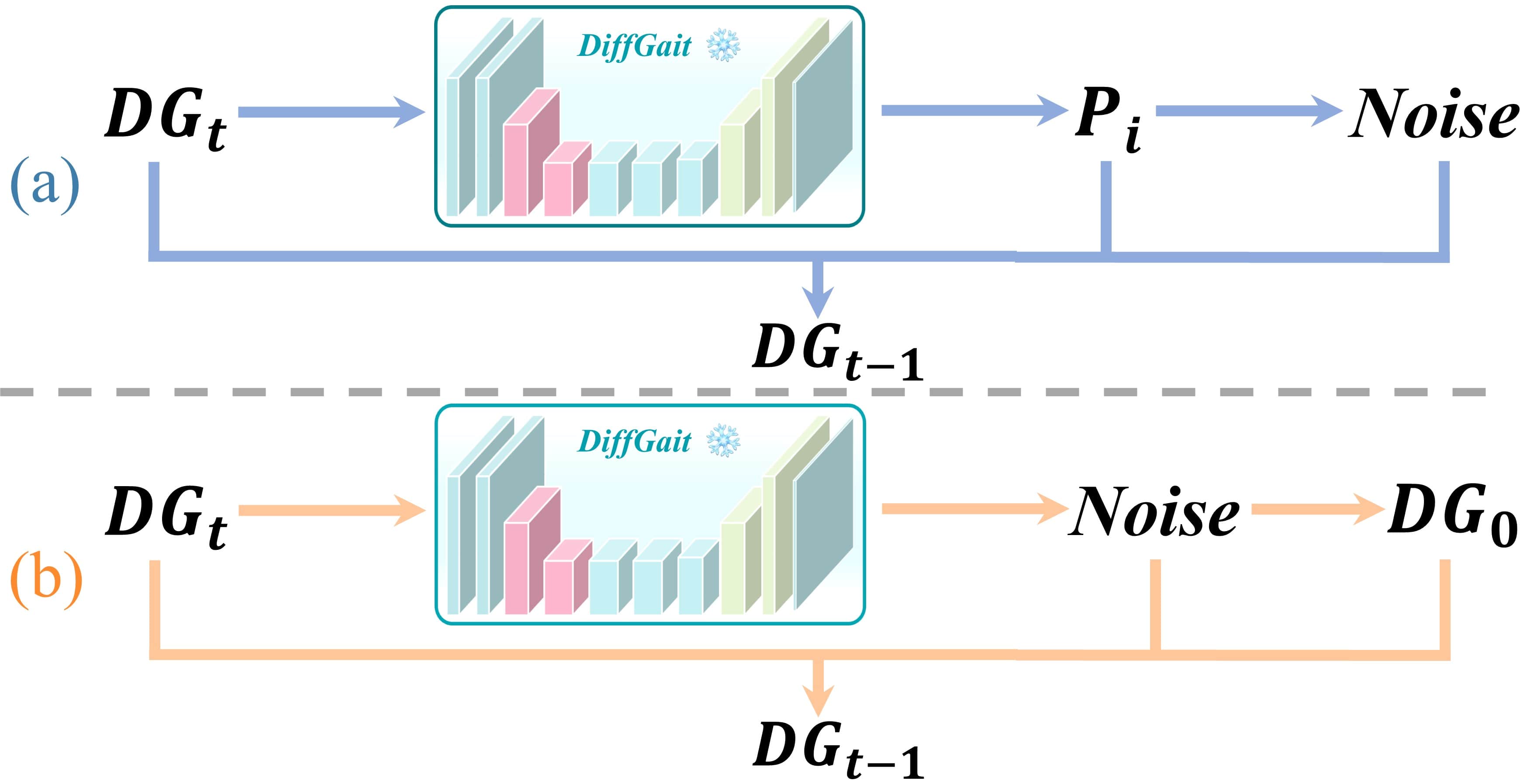}
    \caption{\small Visualization of the reverse process of (a) ours and (b) previous method.}
    \vspace{-6mm}
    \label{diffgait-reverse-process}
\end{figure}

\begin{table*}[t]
    \centering
    \setlength{\abovecaptionskip}{2pt}
    \setlength{\belowcaptionskip}{2pt}
    \caption{\small Comparison of different datasets with their implementation details and setting.}
    \label{tab:dataset}
    \footnotesize
    \begin{tabular}{c|ccc|cccccc}
        \toprule
        \multirow{2}{*}{DataSet} & \multirow{2}{*}{Batch Size} & \multirow{2}{*}{Milestones} & \multirow{2}{*}{Total Steps} & \multicolumn{2}{c}{Train Set} & \multicolumn{2}{c}{Test Set} & \multirow{2}{*}{Scenario} & \multirow{2}{*}{Modalities} \\
         & & & & \#ID & \#Seq & \#ID & \#Seq & & \\
        \hline
        CASIA-B & (8, 16) & (20k, 40k, 50k) & 60k & 74 & 8,140 & 50 & 5,500 & Constrained & Sil., RGB \\
        OU-MVLP & (32, 8) & (60k, 80k, 100k) & 120k & 5,153 & 144,284 & 5,154 & 144,412 & Constrained & Sil., Ske. \\
        Gait3D & (32, 4) & (20k, 40k, 50k) & 60k & 3,000 & 18,940 & 1,000 & 6,369  & Real-world & Sil., Ske., Mesh \\
        GREW & (32, 4) & (80k, 120k, 150k) & 180k & 20,000 & 102,887 & 6,000 & 24,000 & Real-world & Sil., Ske. \\     
        \bottomrule 
    \end{tabular}
    \vspace{-0.3cm}
\end{table*}

\begin{table*}[htbp]
    \setlength{\abovecaptionskip}{2pt}
    \setlength{\belowcaptionskip}{2pt}
    \caption{\small Quantitative evaluation.
    Comparison with other SOTA gait recognition methods across four authoritative datasets. 
    The best performances are in \textbf{blod}, and the second best methods are  \underline{underlined}.}      
    \footnotesize
    \centering
    \label{tab:within-domain}
    \vspace{1mm}
    \begin{tabular}{c|c|cccccccccc}
        \toprule
        \multirow{4}{*}{Input} & \multirow{4}{*}{Method} & \multicolumn{10}{c}{Testing Datasets} \\
        \cline{3-12} & & \multicolumn{4}{c}{\multirow{2}{*}{Gait3D}} & {\multirow{2}{*}{GREW}} & \multirow{2}{*}{OU-MVLP} & \multicolumn{4}{c}{CASIA-B} \\
        \cline{9-12} & & \multicolumn{4}{c}{} & & & NM & BG & CL & Mean \\
        \cline{3-12} & & Rank-1 & Rank-5 & mAP & mINP & Rank-1 & Rank-1 & \multicolumn{4}{c}{Rank-1} \\
        \hline
        \multirow{4}{*}{Silhouette} & GaitSet(AAAI19) & 36.7 & 59.3 & 30.0 & 17.3 & 46.3 & 87.1 & 95.0 & 87.2 & 70.4 & 84.2 \\
                                    & GaitPart(CVPR20) & 28.2 & 47.6 & 21.6 & 12.4 & 44.0 & 88.5 & 96.2 & 91.5 & 78.7 & 88.8 \\
                                    & MTSGait(ACM MM22) & 48.7 & 67.1 & 37.6 & 22.0 & 55.3 & - & - & - & - & - \\
                                    & GaitBase(CVPR23) & 64.6 & - & - & - & 60.1 & 90.8 & 97.6 & 94.0 & 77.4 & 89.7 \\
        \hline
        \multirow{5}{*}{Skeleton} & GaitGraph(ICIP21) & 12.6 & 28.7 & 11.0 & 6.5 & 10.2 & 4.2 & 86.3 & 76.5 & 65.2 & 76.0 \\
                                & GaitGraph2(CVPRW22) & 7.2 & 15.9  & 5.2 & 3.0 & 34.8 & 70.7 & 80.3 & 71.4 & 63.8 & 71.8 \\
                                & GPGait(ICCV23) & 22.4 & - & - & - & 57.0 & 59.1 & 93.6 & 80.2 & 69.3 & 81.0 \\
                                & SkeletonGait(AAAI24) & \underline{38.1} & \underline{56.7} & \underline{28.9} & \underline{16.1} & \textbf{77.4} & \underline{67.4} & - & - & - & - \\
                                \rowcolor{yellow!20} & Ours & \textbf{39.5} & \textbf{60.1} & \textbf{30.4} & \textbf{17.1} & \underline{72.5} & \textbf{68.2} & \textbf{94.1} & \textbf{81.3} & \textbf{74.9} & \textbf{83.4} \\
        \bottomrule 
    \end{tabular}
    \vspace{-0.3cm}
\end{table*}

\noindent \textbf{DiffGait Reverse Process.}
We have modified the DDIM \cite{16-song2020denoising} sampling method to generate multi-level silhouettes under different timesteps and reduce Gaussian uncertainty, as shown in Fig.~\ref{diffgait-reverse-process}. 

Standard Gaussian noise $DG_T $ serves as the initial input. 
Then, $HGV_T$ is calculated based on Equation~\ref{hgv}, and $P_T$ is predicted by $\mathcal{DE}$. 
We consider $P_T$ as the maximum likelihood of $DG_0$, denoted as $DG_{0}^{T}$. 
Then we calculate the noise for the current timestep via the following formula:
\begin{equation}
    \epsilon=\frac{1}{\sqrt{1-\overline{\alpha}_{T}}}(DG_{T}-\sqrt{\overline{\alpha}_{T}}DG_{T}^{0}).
\end{equation}
Furthermore, $\epsilon$ and $DG_{0}^{l}$ are used to recover the $DG_{T-1}$. The formulation is expressed as,
\begin{equation}
    DG_{T-1}=\sqrt{\overline{\alpha}_{T-1}}DG_{T}^{0} + \sigma\epsilon^{*} + \sqrt{1-\overline{\alpha}_{T-1}-\sigma^{2}}\epsilon,
\end{equation}
where $\sigma=\eta\sqrt{(1-\frac{\overline{\alpha}_{T}}{\overline{\alpha}_{T-1}})\cdot\frac{1-\overline{\alpha}_{T-1}}{1-\overline{\alpha}_{T}}}$, $\eta$ means sampling coefficient and $\epsilon^{*}$ means the Standard Gaussian Noise.
$DG_{T-1}$ will continue to participate in the reverse process, following the method mentioned above, until the end of the sampling process.

\noindent \textbf{DiffGait Architecture.}
The inference speed and model parameter size exhibited by existing UNet-based diffusion models render their application to gait recognition entirely impractical.
Considering gait modalities are 'clean', DiffGait employs an extremely streamlined architecture with a surprisingly low model parameter size of 1.9M and an impressive inference speed of 3628 FPS.

Compared with previous models, DiffGait incorporates two key improvements:
\textbf{1)} DiffGait adopts a Decoder-only architecture. 
\textbf{2)} DiffGait eliminates unnecessary convolutional layers and attention modules, operating within a low-dimensional feature space.
To be specific, a five-layer ResNet-like encoder $\mathcal{E}$ is designed to extract skeleton structure features $G_{ske}$.
\textit{Gait Mapping} is adopted for downsampling silhouette-based Gaussian noise $DG_t$.
A similarly sized decoder $\mathcal{DE}$, denoises silhouettes from Hybrid Gait Volume (HGV) integrated by $G_{ske}$ and $DG_t$.

\subsection{3.4 Perceptual Gait Integration (PGI)}
Through the above designs, DiffGait produces multi-level silhouettes $\{ P_i \}_{i=1}^{M}$, each emphasizing various appearance feature levels.
Effective integration of structure and appearance features to construct a comprehensive gait representation significantly influences the overall recognition performance, so we propose Perceptual Gait Integration (PGI) consisting of two stages:

\textbf{In stage one}, our strategy is to allocate varying weights to different timesteps to generate the well-defined silhouette $\mathcal{P}_t$, acknowledging that the reconstruction process progresses from local to global features, with later steps yielding enhanced outcomes.
We set the weights $\omega$ as (0, 0, 0.2, 0.3, 0.5) and discuss them in Tab.~\ref{alb: weights}.

\textbf{In stage two}, our goal is to mitigate noise arising from gait modalities fusion and effectively integrate structure and appearance features.
The initialized $\mathcal{P}_t$ and $\mathcal{I}_t$ are concatenated to create a fused feature, which serves as the input for subsequent processing.
Following \cite{63-Min2024GaitMAPM}, by employing a perceptual structure centered on ReLU and Sigmoid activations, we assign higher weights to features that are similar across skeletons and silhouettes, eliminate the impact of redundant features, and derive $\mathcal{H}_t$ through an attention-like approach \cite{37-vaswani2017attention}.

\section{4 Experiment}

\begin{figure*}[t]
    \setlength{\abovecaptionskip}{2pt}
    \centering
    \includegraphics[width=0.9\linewidth]{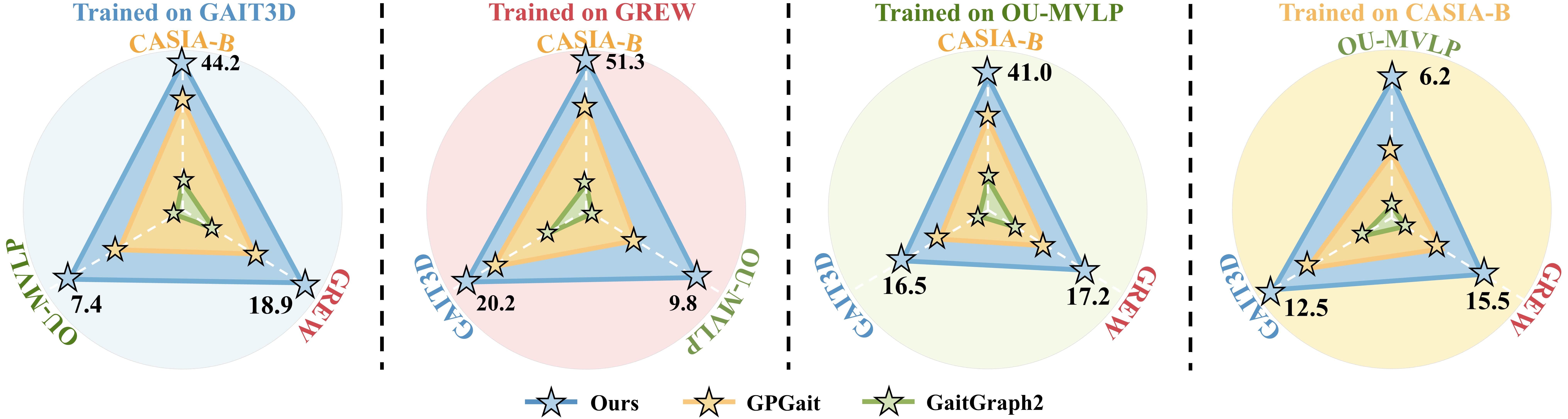}
    \caption{
    \small Comparison with other methods on four authoritative Datasets. 
    The proposed method achieves state-of-the-art generalization performance across in all scenarios.
    }
    \label{cross}
    \vspace{-0.4cm}
\end{figure*}

\subsection{4.1 Experimental Settings}
\noindent \textbf{Datasets \& Metrics.}
We evaluated our proposed method on four mainstream datasets, including two outdoor datasets: Gait3D \cite{21-zheng2022gait} GREW \cite{28-zhu2021gait}, and two indoor datasets: CASIA-B \cite{26-yu2006framework} OU-MVLP \cite{27-takemura2018multi}.
The key statistics of these gait datasets are listed in Tab.~\ref{tab:dataset}.
We use the following metrics to evaluate model performance quantitatively: rank retrieval (Rank-1, Rank-5), mean Average Precision(mAP) and mean Inverse Negative Penalty (mINP), and Rank-1 accuracy is considered the primary metric.

\noindent \textbf{Implementation details.}
During the training and inference stages, we use PyTorch as the framework to conduct all experiments on two RTX 3090. 
For the full four datasets, we used silhouettes with a resolution of $64 \times 44$, and skeletons are the 2D coordinates of joints that conform to COCO 17 and transformed into Heat-skeletons with a resolution of $2 \times 64 \times 44$.
Tab.~\ref{tab:dataset} displays the main hyper-parameters of our experiments.

When training DiffGait, we choose Adam as the optimizer, with the initial learning rate set to 0.01 and batch size (16, 4). We set the timestep as 1000. A cosine schedule is adopted to set the noise coefficient $\beta_{t}$, $\alpha_{t}$. 
When training ZipGait, the SGD optimizer with an initial learning rate of 0.1 and weight decay of 0.0005 is utilized.

\subsection{4.2 Comparison with State-of-the-art Methods}
\subsubsection{4.2.1 Quantitative comparison}
We initiate our analysis with an \textit{Intra-domain comparison}, where we systematically evaluate our approach against the current state-of-the-art model-based methods, with comparative results presented in Tab.~\ref{tab:within-domain}. 
Our method improves the performance by 10.2\% compared to GPGait and 1.8\% compared to SkeletonGait in Gait3D.
Similarly, our method also holds a significant advantage over other methods in the indoor datasets.

We further perform a \textit{Cross-domain comparison} to assess our method's efficacy across different testing scenarios, and detailed results are shown in Fig.~\ref{cross}.
We compare two representative skeleton-based methods, GaitGraph2 and GPGait, with the latter demonstrating superior generalization capabilities.
We can observe from the experimental results: 
1) Skeletons are susceptible to data distribution; 
2) Reconstructed appearance features can enhance representation robustness;
3) Training on real datasets leads to improved generalization capabilities.

\begin{figure}[t]
    \setlength{\abovecaptionskip}{2pt}
    \setlength{\belowcaptionskip}{2pt}
    \centering
    \includegraphics[width=0.95\linewidth]{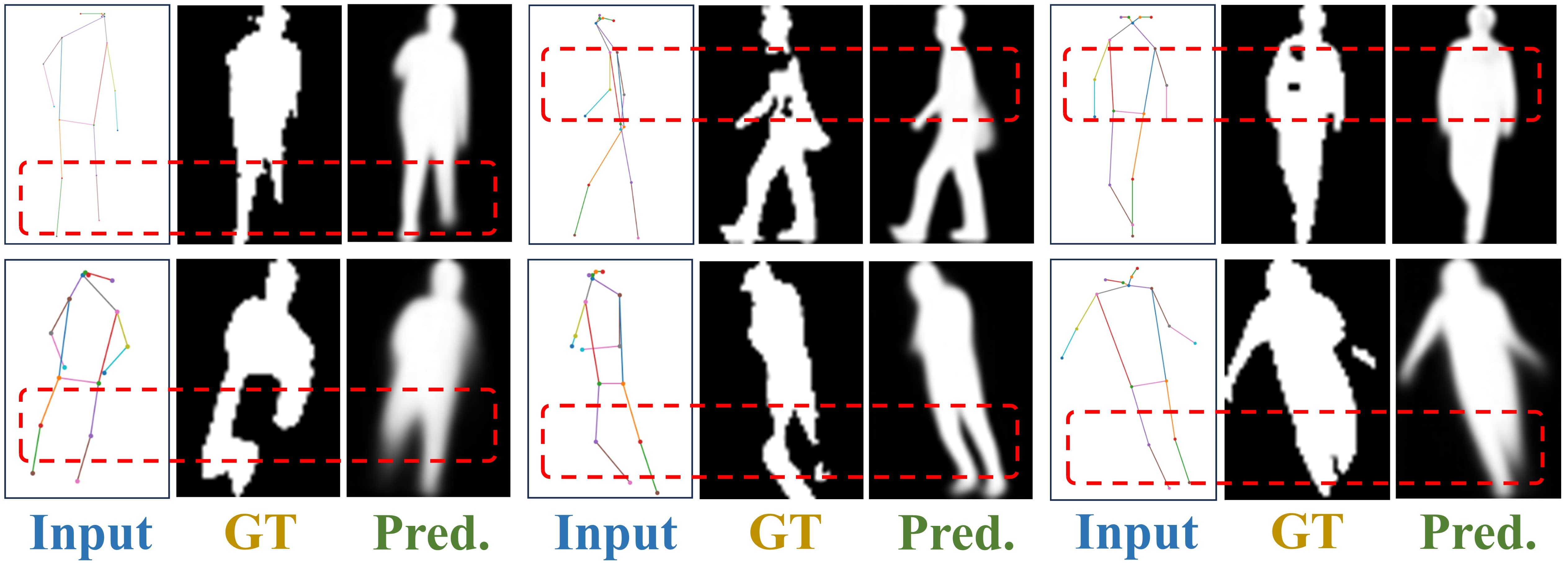}
    \caption{\small Qualitative evaluation. 
    Visualization of the silhouette reconstruction results of DiffGait in diverse environments. 
    Red boxes highlight the regions where features are missing.}
    \label{visual_analysis}
    \vspace{-0.4cm}
\end{figure}

\begin{table}[!t]
    \setlength{\abovecaptionskip}{2pt}
    \setlength{\belowcaptionskip}{2pt}
    \caption{\small Efficiency comparison with different metrics.
    All models are evaluated on a single 3090 GPU.
    Params represents the number of model parameters.}
    \footnotesize
    \label{diffusion-model-comparison}
    \centering
    \begin{tabular}{c|c|c}
        \toprule
        Model & Inference speed (FPS) $\uparrow$  & Params (M) $\downarrow$ \\
        \midrule
        DDPM & 345 & 45.09 \\
        DDIM & 1754 & 45.09 \\
        \rowcolor{yellow!20} DiffGait (ours) & \textbf{3628} & \textbf{1.9} \\
        \bottomrule
    \end{tabular}
    \vspace{-0.5cm}
\end{table}

\subsubsection{4.2.2 Qualitative comparison}
As illustrated in Fig.~\ref{visual_analysis}, we randomly selected over a dozen test skeletons from four datasets.
DiffGait can efficiently reconstruct occluded body parts by learning the a priori distribution of the human body. 
It can be observed that, compared to GT, various gait noises, such as occlusions, have been eliminated.

\begin{figure}[t]
    \setlength{\abovecaptionskip}{2pt}
    \setlength{\belowcaptionskip}{2pt}
    \centering
    \includegraphics[width=0.95\linewidth]{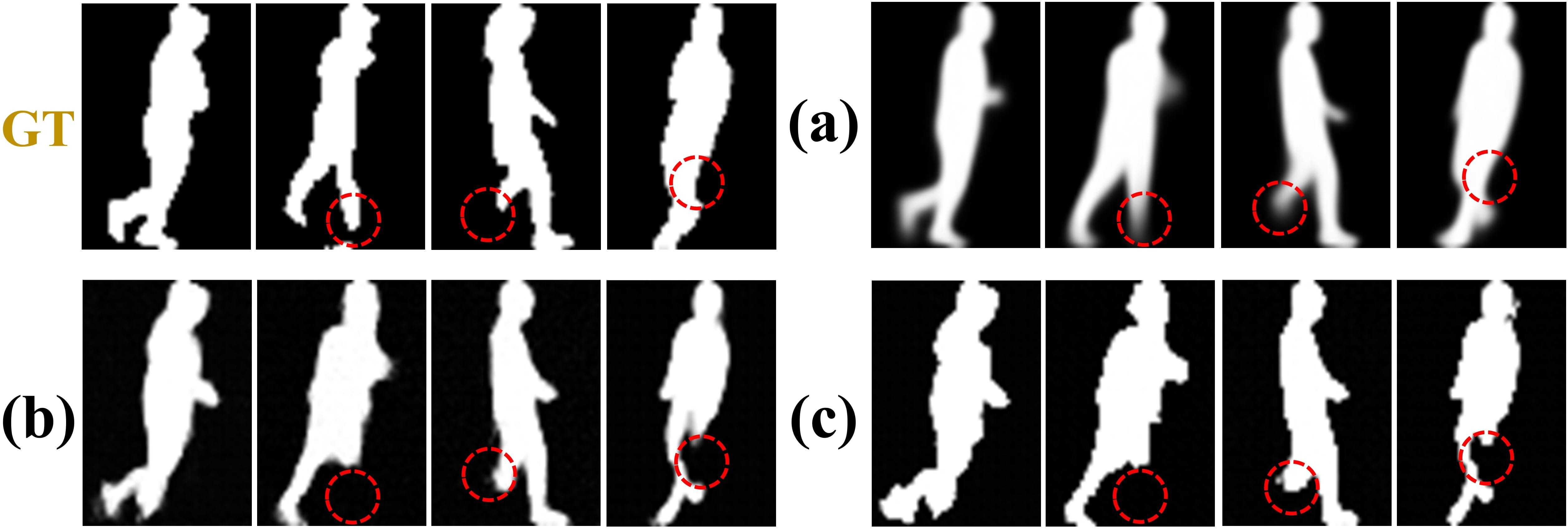}
    \caption{\small Qualitative evaluation. 
    Visualization of the silhouette reconstruction results of our DiffGait (a), DDIM (b), DDPM (c) in diverse environments. Red circles mean challenging regions.
}
    \label{visual_comparison}
    \vspace{-0.4cm}
\end{figure}

We further examine the ability of our model in dealing with challenging scenarios. 
We depict in Fig.~\ref{visual_comparison} side-by-side comparisons of (a) our DiffGait against standard (b) DDIM and (c) DDPM, GT stands for ground truth.
It is observed that our DiffGait consistently reconstructs silhouettes for various challenging scenes.

\subsubsection{4.2.3 Efficiency comparison.}
In addition, to demonstrate the efficiency of our proposed method, we compare the number of model parameters and inference speed between our DiffGait and two typical methods DDPM and DDIM in Tab.~\ref{diffusion-model-comparison}.
DiffGait achieves significant performance while only requiring approximately 1.9M model parameters. 
It outperforms them by achieving a speedup of about 10 times and 2 times faster in terms of Frames Per Second (FPS).

\begin{table}[t]
    \setlength{\abovecaptionskip}{2pt}
    \setlength{\belowcaptionskip}{2pt}
    \caption{\small Evaluation of Plug-and-Play Performance of DiffGait.}
    \label{plug-and-play performance: diffgait}
    \footnotesize
    \centering
    \begin{tabular}{l|ll}
        \toprule
        \diagbox{Methods}{Metrics} & Rank-1 (\%) & mAP (\%) \\
        \midrule
        GaitGraph(ICIP21) & 12.6 & 11.0 \\
        GaitGraph + DiffGait & 23.2 (10.6 $\uparrow$) & 17.1 (6.1 $\uparrow$) \\       
        \midrule
        GaitTR(ES23) & 7.8 & 6.6 \\
        GaitTR + DiffGait & 19.6 (11.8 $\uparrow$) & 15.7 (9.1 $\uparrow$) \\
        \midrule
        GaitGraph2(CVPRW22) & 7.2 & 5.2 \\
        GaitGraph2 + DiffGait & 18.9 (11.7 $\uparrow$) & 15.9 (10.7 $\uparrow$) \\
        \midrule
        GPGait(ICCV23) & 22.3 & 16.5 \\
        GPGait + DiffGait & 31.3 (9.0 $\uparrow$) & 22.9 (6.4 $\uparrow$) \\   
        \midrule
        SkeletonGait(AAAI24) & 38.1 & 28.9 \\
        SkeletonGait + DiffGait & 39.2 (1.1 $\uparrow$) & 29.5 (0.6 $\uparrow$) \\
        \bottomrule
    \end{tabular}
    \vspace{-0.4cm}
\end{table}

\subsection{4.3 Plug-and-Play Performance Evaluation.}
we demonstrate the plug-and-play performance of DiffGait by applying it to state-of-the-art model-based methods \cite{6-teepe2021gaitgraph, 7-teepe2022towards, 28-zhang2023spatial, 8-fu2023gpgait, 13-fan2024skeletongait}.
The experimental outcomes, detailed in Tab.~\ref{plug-and-play performance: diffgait}, reveal that DiffGait significantly enhances the performance of all existing model-based methods across various metrics. 
The effectiveness stems from reconstructed silhouettes, which provides richer gait features. 
Contrasting with ZipGait, we integrate gait features by padding and concatenating before passing them into the head layer. 

\begin{figure}[!t]
    \setlength{\abovecaptionskip}{2pt}
    \setlength{\belowcaptionskip}{2pt}
    \centering
    \includegraphics[width=0.95\linewidth]{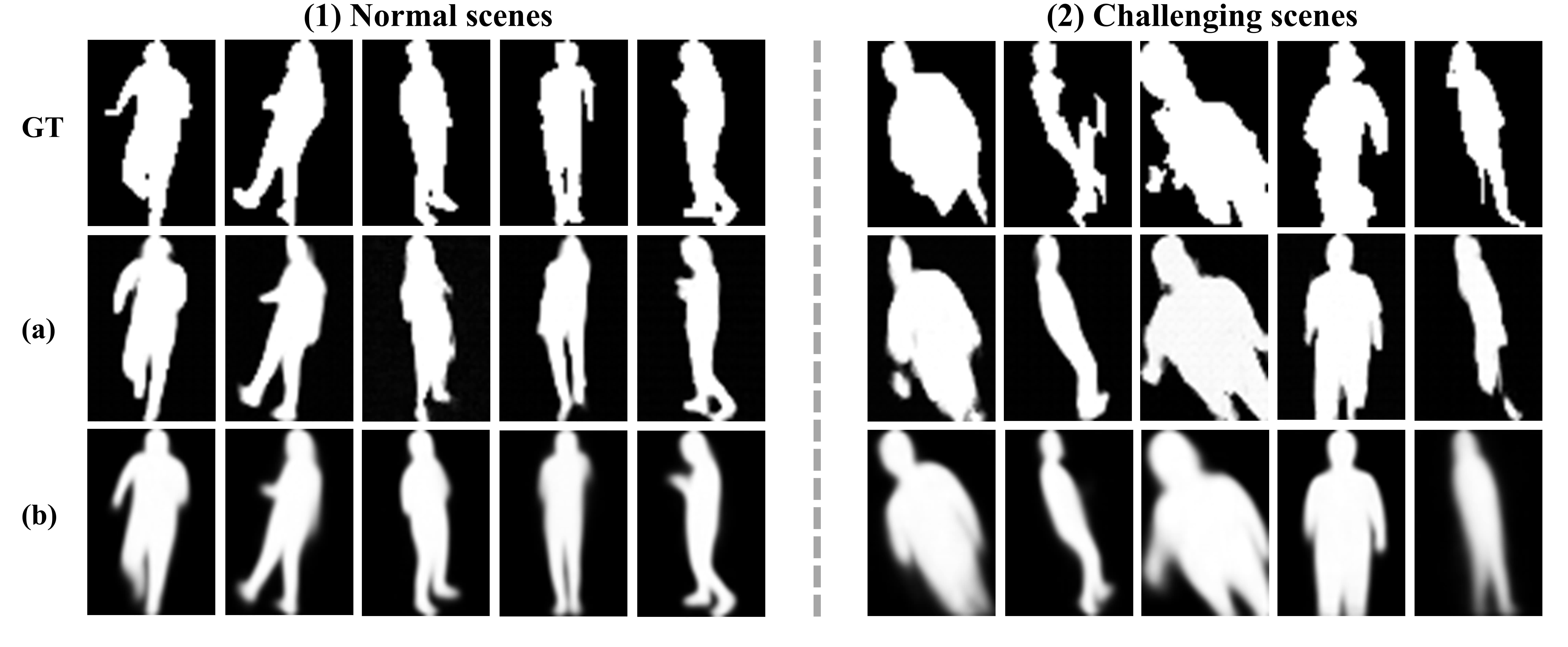}
    \caption{\small Visualization of the reconstructed results in normal and challenging scenarios. (a) donates DDIM, (b) donates DiffGait.}
    \label{abl:different_scenes}
    \vspace{-0.7cm}
\end{figure}

\subsection{4.4 Ablation Study}

\noindent \textbf{Study on components of ZipGait.}
We empirically evaluate the effectiveness of each proposed component, and report the results in Tab.~\ref{ablation: structure}
(a) is our baseline, using Heat-skeletons as input;
(b) solely uses silhouettes predicted by DiffGait;
(c) illustrates a simple sum fusion of shape and structure features;
(d) demonstrates that our proposed PGI improves performance from 38.7\% to 39.5\%.
\begin{table}[h]
    \vspace{-0.3cm}
    \setlength{\abovecaptionskip}{2pt}
    \setlength{\belowcaptionskip}{2pt}
    \caption{\small Ablation of different components in ZipGait on Gait3D.
    Heat. denotes Heat-skeleton; Silh. donates Silhouette; PGI donates Perceptual Gait Integration.}
    \label{ablation: structure}
    \footnotesize
    \centering
    \begin{tabular}{l|ccc|cc}
        \toprule
        Structure & Heat.& Silh. & PGI & Rank-1 & mAP \\
        \hline
        (a) Baseline & \checkmark & \ding{55} & \ding{55} & 33.6 & 23.4 \\
        (b) & \ding{55} & $\checkmark$ & \ding{55} & 24.7 & 18.1\\
        (c) & $\checkmark$ & $\checkmark$ & \ding{55} & 38.7 & 27.2 \\
        \rowcolor{yellow!20} (d) ZipGait & $\checkmark$ & $\checkmark$ & $\checkmark$ & \textbf{39.5} & \textbf{30.4} \\
        \bottomrule
    \end{tabular}
    \vspace{-0.7cm}
\end{table}
\begin{table}[h]
    \setlength{\abovecaptionskip}{2pt}
    \setlength{\belowcaptionskip}{2pt}
    \caption{\small Ablation of various designs in PGI on Gait3D.}
    \label{ablation: PGI}
    \footnotesize
    \centering
    \begin{tabular}{l|cc|cc}
        \toprule
        Structure & Stage 1 & Stage 2 & Rank-1 & mAP \\
        \hline
        (a) & $\checkmark$ & \ding{55} & 39.0 & 28.9 \\
        (b) & \ding{55} & $\checkmark$ & 39.2 & 29.5\\
        \rowcolor{yellow!20} (c) Full & $\checkmark$ & $\checkmark$ & \textbf{39.5} & \textbf{30.4} \\
        \bottomrule
    \end{tabular}
    \vspace{-0.3cm}
\end{table}

\noindent \textbf{Study on Perceptual Gait Integration.}
We further explore the influence of two stages within PGI, and tabulate the results in Tab.~\ref{ablation: PGI}.
(a) only utilizes stage 1, refining the predicted multi-level silhouettes through dynamic weight allocation;
(b) employs a gait fusion layer to generate mixed gait features, representing stage 2;
(c) demonstrates the complete two-stage integration.
Each stage independently enhances performance, and their integration results in even greater improvements, thereby proving the efficacy of the modules.
\begin{table}[H]
    \vspace{-0.4cm}
    \setlength{\abovecaptionskip}{2pt}
    \setlength{\belowcaptionskip}{2pt}
    \caption{\small Ablation of different diffusion process.}
    \label{ablation: Process}
    \footnotesize
    \centering
    \begin{tabular}{l|cccc}
        \toprule
        Method & Rank-1 & Rank-5 & mAP & mINP\\
        \hline
         DDIM Process& 38.4 & 58.1 & 29.2 & 16.5 \\
        \rowcolor{yellow!20} DiffGait Process & \textbf{39.5} & \textbf{60.1} & \textbf{30.4}  & \textbf{17.1} \\
        \bottomrule
    \end{tabular}
    \vspace{-0.4cm}
\end{table}

\begin{figure}[!t]
    \setlength{\abovecaptionskip}{2pt}
    \setlength{\belowcaptionskip}{2pt}
    \centering
    \includegraphics[width=0.95\linewidth]{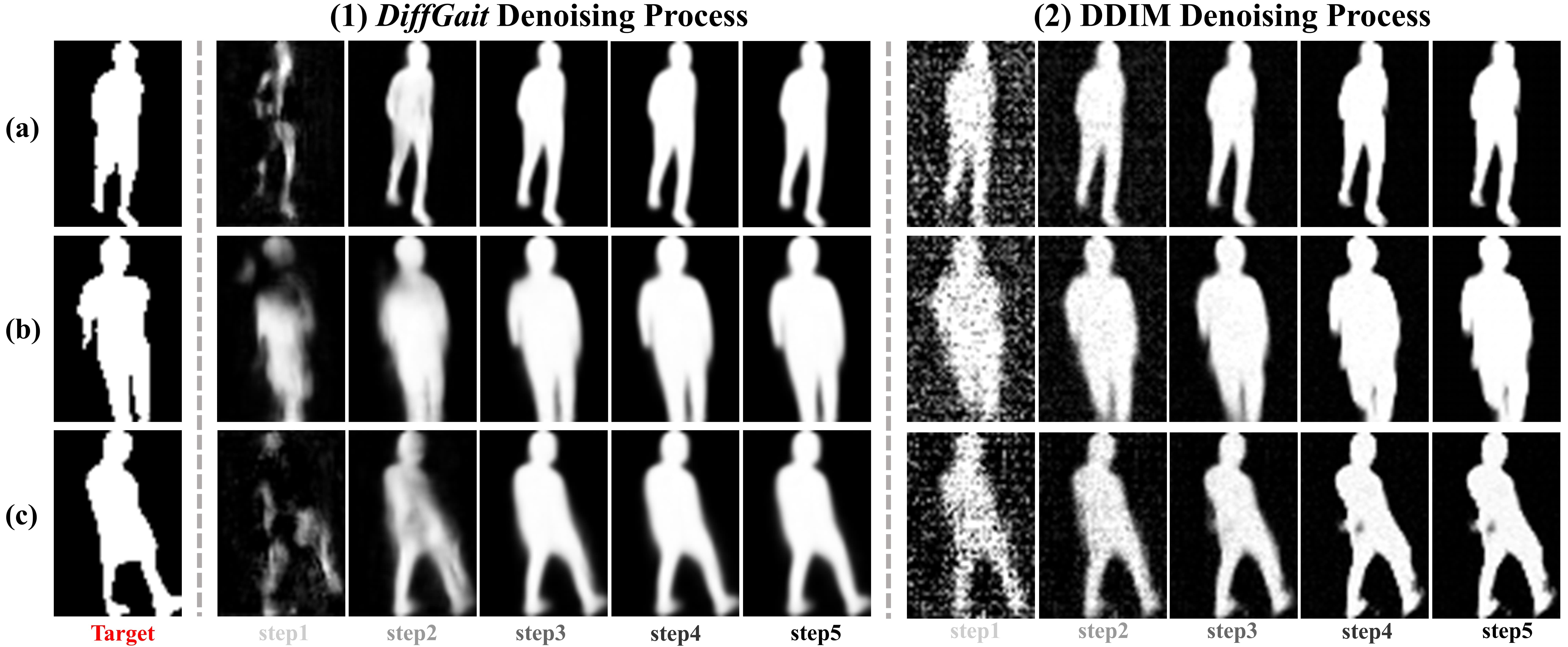}
    \caption{\small Visualization of the reconstructed results about whole denoising process under different algorithms.}
    \label{abl:denoising_process}
    \vspace{-0.4cm}
\end{figure}

\noindent \textbf{Study on DiffGait Process.}
To validate the effect of the DiffGait Process, we consider two alternative ways to train our DiffGait.
Fig.~\ref{abl:denoising_process} visualizes the denoising process in detail, highlighting the distinctions between the two methods.
Further experiments under identical conditions demonstrated superior performance of the DiffGait Process, providing a clear validation of its effectiveness, shown in Tab.~\ref{ablation: Process}.
Model training and inference of DDIM process can be found in Appendix, along with more detailed comparisons.

\begin{figure}[!t]
    \setlength{\abovecaptionskip}{2pt}
    \setlength{\belowcaptionskip}{2pt}
    \centering
    \includegraphics[width=1\linewidth]{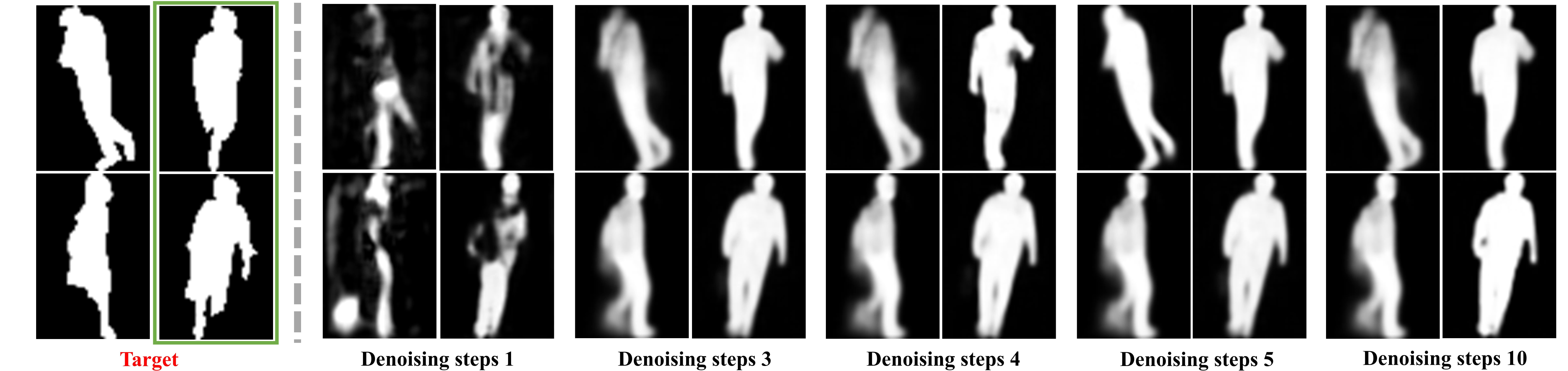}
    \caption{\small Visualization of the reconstructed results under different denoising steps of DiffGait.}
    \label{abl:sampling_steps}
    \vspace{-0.7cm}
\end{figure}

\subsection{4.5 Further Analysis}

\noindent \textbf{Denoising steps analysis.}
As shown in Fig.~\ref{abl:sampling_steps}, setting the denoising steps to three or more can generate effective silhouettes. 
To select the most appropriate number of denoising steps, we analyzed several possible values, with performance and efficiency serving as the criteria. 
The experimental results in Tab.~\ref{alb: sampling steps} reveal that the best performance is achieved with \textbf{5} denoising steps, striking a balance by enhancing performance without significant additional resource demands, with only a slight 0.2s increase per iteration.
\begin{table}[H]
    \vspace{-0.4cm}
    \setlength{\abovecaptionskip}{2pt}
    \setlength{\belowcaptionskip}{2pt}
    \caption{\small Parameter analysis. 
    Comparison model performance on Gait3D under different sampling steps for DiffGait.}
    \footnotesize
    \label{alb: sampling steps}
    \centering
    \begin{tabular}{c|ccc}
        \toprule
        Sampling Steps & Training time  & Rank-1 & mAP \\
        \midrule
        No DiffGait & 37.2s & 33.6 & 23.4 \\
        1 sampling steps & 43.8s & 6.8 & 1.7 \\
        3 sampling steps & 53.4s & 37.4 & 27.1 \\
        4 sampling steps & 54.2s & 38.4 & 28.3 \\
        \rowcolor{yellow!20} 5 sampling steps & 59.6s & \textbf{38.7} & \textbf{29.2} \\
        10 sampling steps & 76.4s & 37.9 & 28.8 \\    
        \bottomrule
    \end{tabular}
    \vspace{-0.4cm}
\end{table}

\noindent \textbf{Dynamic weight allocation analysis.}
As shown in Fig.~\ref{abl:denoising_process}, during the denoising process, the quality of predicted silhouettes varies at each step.
We established various potential combinations of weights and conducted experimental validations. 
The results, displayed in Tab.~\ref{alb: weights}, indicate that $\alpha_4$ is the optimal choice for Stage 1 of PGI.
\begin{table}[H]
    \vspace{-0.4cm}
    \setlength{\abovecaptionskip}{2pt}
    \setlength{\belowcaptionskip}{2pt}
    \caption{\small Parameter analysis. 
    Comparison model performance on Gait3D datasets under different weight combinations.}
    \label{alb: weights}
    \footnotesize
    \centering
    \begin{tabular}{c|ccccc|cc}
        \toprule
        groups & $\omega_{1}$ & $\omega_{2}$ & $\omega_{3}$ & $\omega_{4}$ & $\omega_{5}$ & Rank-1 & mAP \\
        \hline
        $\alpha_{1}$ & 1 & 0 & 0 & 0 & 0 & 12.8 & 6.2 \\
        $\alpha_{2}$ & 0 & 0 & 0 & 0 & 1 & 39.0 & 29.4 \\
        $\alpha_{3}$ & 0 & 0 & 0 & 0.5 & 0.5 & 39.2 & 29.8 \\
        \rowcolor{yellow!20} $\alpha_{4}$ & 0 & 0 & 0.2 & 0.3 & 0.5 & \textbf{39.5} & \textbf{30.4} \\
        $\alpha_{5}$ & 0.2 & 0.2 & 0.2 & 0.2 & 0.2 & 38.7 & 29.2 \\
        \bottomrule
    \end{tabular}
    \vspace{-0.4cm}
\end{table}

\section{5 Conclusion}
This paper establishes the connection between skeletons and silhouettes via the diffusion model, providing new insights into utilizing gait modalities.
The proposed DiffGait is the first successful method to reconstruct dense body shapes from sparse skeleton structures.
Simultaneously, Perceptual Gait Integration is proposed for efficient fusion of gait modality.
Ultimately, ZipGait demonstrates superior performance in both cross-domain and intra-domain settings and achieves plug-and-play improvements.
This study highlights the potential of gait modality interaction in gait recognition.
\bibliography{aaai25.bib}

\begin{thebibliography}{57}
\providecommand{\natexlab}[1]{#1}

\bibitem[{Chao et~al.(2019)Chao, He, Zhang, and Feng}]{3-chao2019gaitset}
Chao, H.; He, Y.; Zhang, J.; and Feng, J. 2019.
\newblock Gaitset: Regarding gait as a set for cross-view gait recognition.
\newblock In \emph{Proceedings of the AAAI conference on artificial intelligence}, volume~33, 8126--8133.

\bibitem[{Chen et~al.(2023)Chen, Sun, Song, and Luo}]{48-chen2023diffusiondet}
Chen, S.; Sun, P.; Song, Y.; and Luo, P. 2023.
\newblock Diffusiondet: Diffusion model for object detection.
\newblock In \emph{Proceedings of the IEEE/CVF international conference on computer vision}, 19830--19843.

\bibitem[{Choi et~al.(2019)Choi, Kim, Kim, and Kim}]{30-choi2019skeleton}
Choi, S.; Kim, J.; Kim, W.; and Kim, C. 2019.
\newblock Skeleton-based gait recognition via robust frame-level matching.
\newblock \emph{IEEE Transactions on information forensics and security}, 14(10): 2577--2592.

\bibitem[{Cui and Kang(2023)}]{19-cui2023multi}
Cui, Y.; and Kang, Y. 2023.
\newblock Multi-modal gait recognition via effective spatial-temporal feature fusion.
\newblock In \emph{Proceedings of the IEEE/CVF conference on computer vision and pattern recognition}, 17949--17957.

\bibitem[{Dong et~al.(2024)Dong, Yu, Ha, Shi, Ma, Xu, Fu, and Wang}]{20-dong2024hybridgait}
Dong, Y.; Yu, C.; Ha, R.; Shi, Y.; Ma, Y.; Xu, L.; Fu, Y.; and Wang, J. 2024.
\newblock HybridGait: A benchmark for spatial-temporal cloth-changing gait recognition with hybrid explorations.
\newblock In \emph{Proceedings of the AAAI conference on artificial intelligence}, volume~38, 1600--1608.

\bibitem[{Dou et~al.(2023)Dou, Zhang, Su, Yu, Lin, and Li}]{43-dou2023gaitgci}
Dou, H.; Zhang, P.; Su, W.; Yu, Y.; Lin, Y.; and Li, X. 2023.
\newblock Gaitgci: Generative counterfactual intervention for gait recognition.
\newblock In \emph{Proceedings of the IEEE/CVF conference on computer vision and pattern recognition}, 5578--5588.

\bibitem[{Dou et~al.(2024)Dou, Zhang, Zhao, Jin, and Li}]{39-dou2024clash}
Dou, H.; Zhang, P.; Zhao, Y.; Jin, L.; and Li, X. 2024.
\newblock CLASH: Complementary Learning with Neural Architecture Search for Gait Recognition.
\newblock \emph{IEEE Transactions on image processing}.

\bibitem[{Duan et~al.(2022)Duan, Zhao, Chen, Lin, and Dai}]{25-duan2022revisiting}
Duan, H.; Zhao, Y.; Chen, K.; Lin, D.; and Dai, B. 2022.
\newblock Revisiting skeleton-based action recognition.
\newblock In \emph{Proceedings of the IEEE/CVF conference on computer vision and pattern recognition}, 2969--2978.

\bibitem[{Fan et~al.(2023)Fan, Liang, Shen, Hou, Huang, and Yu}]{4-fan2023opengait}
Fan, C.; Liang, J.; Shen, C.; Hou, S.; Huang, Y.; and Yu, S. 2023.
\newblock Opengait: Revisiting gait recognition towards better practicality.
\newblock In \emph{Proceedings of the IEEE/CVF conference on computer vision and pattern recognition}, 9707--9716.

\bibitem[{Fan et~al.(2024)Fan, Ma, Jin, Shen, and Yu}]{13-fan2024skeletongait}
Fan, C.; Ma, J.; Jin, D.; Shen, C.; and Yu, S. 2024.
\newblock SkeletonGait: Gait Recognition Using Skeleton Maps.
\newblock In \emph{Proceedings of the AAAI conference on artificial intelligence}, volume~38, 1662--1669.

\bibitem[{Fan et~al.(2020)Fan, Peng, Cao, Liu, Hou, Chi, Huang, Li, and He}]{5-fan2020gaitpart}
Fan, C.; Peng, Y.; Cao, C.; Liu, X.; Hou, S.; Chi, J.; Huang, Y.; Li, Q.; and He, Z. 2020.
\newblock Gaitpart: Temporal part-based model for gait recognition.
\newblock In \emph{Proceedings of the IEEE/CVF conference on computer vision and pattern recognition}, 14225--14233.

\bibitem[{Feng et~al.(2023)Feng, Gao, Tse, Ma, and Chang}]{47-feng2023diffpose}
Feng, R.; Gao, Y.; Tse, T. H.~E.; Ma, X.; and Chang, H.~J. 2023.
\newblock DiffPose: SpatioTemporal diffusion model for video-based human pose estimation.
\newblock In \emph{Proceedings of the IEEE/CVF international conference on computer vision}, 14861--14872.

\bibitem[{Frank, Mannor, and Precup(2010)}]{59-frank2010activity}
Frank, J.; Mannor, S.; and Precup, D. 2010.
\newblock Activity and gait recognition with time-delay embeddings.
\newblock In \emph{Proceedings of the AAAI conference on artificial intelligence}, volume~24, 1581--1586.

\bibitem[{Fu et~al.(2023)Fu, Meng, Hou, Hu, and Huang}]{8-fu2023gpgait}
Fu, Y.; Meng, S.; Hou, S.; Hu, X.; and Huang, Y. 2023.
\newblock Gpgait: Generalized pose-based gait recognition.
\newblock In \emph{Proceedings of the IEEE/CVF international conference on computer vision}, 19595--19604.

\bibitem[{Fu et~al.(2019)Fu, Wei, Zhou, Shi, Huang, Wang, Yao, and Huang}]{29-fu2019horizontal}
Fu, Y.; Wei, Y.; Zhou, Y.; Shi, H.; Huang, G.; Wang, X.; Yao, Z.; and Huang, T. 2019.
\newblock Horizontal pyramid matching for person re-identification.
\newblock In \emph{Proceedings of the AAAI conference on artificial intelligence}, volume~33, 8295--8302.

\bibitem[{Gao et~al.(2022)Gao, Yun, Zhao, and Liu}]{58-gao2022gait}
Gao, S.; Yun, J.; Zhao, Y.; and Liu, L. 2022.
\newblock Gait-D: skeleton-based gait feature decomposition for gait recognition.
\newblock \emph{IET computer vision}, 16(2): 111--125.

\bibitem[{Gong et~al.(2023)Gong, Foo, Fan, Ke, Rahmani, and Liu}]{23-gong2023diffpose}
Gong, J.; Foo, L.~G.; Fan, Z.; Ke, Q.; Rahmani, H.; and Liu, J. 2023.
\newblock Diffpose: Toward more reliable 3d pose estimation.
\newblock In \emph{Proceedings of the IEEE/CVF conference on computer vision and pattern recognition}, 13041--13051.

\bibitem[{Guo et~al.(2023)Guo, Shah, Liu, Chellappa, and Peng}]{14-guo2023gaitcontour}
Guo, Y.; Shah, A.; Liu, J.; Chellappa, R.; and Peng, C. 2023.
\newblock GaitContour: Efficient Gait Recognition based on a Contour-Pose Representation.
\newblock \emph{arXiv preprint arXiv:2311.16497}.

\bibitem[{Han and Bhanu(2005)}]{55-han2005individual}
Han, J.; and Bhanu, B. 2005.
\newblock Individual recognition using gait energy image.
\newblock \emph{IEEE Transactions on pattern analysis and machine intelligence}, 28(2): 316--322.

\bibitem[{He et~al.(2016)He, Zhang, Ren, and Sun}]{36-he2016deep}
He, K.; Zhang, X.; Ren, S.; and Sun, J. 2016.
\newblock Deep residual learning for image recognition.
\newblock In \emph{Proceedings of the IEEE conference on computer vision and pattern recognition}, 770--778.

\bibitem[{Hermans, Beyer, and Leibe(2017)}]{38-hermans2017defense}
Hermans, A.; Beyer, L.; and Leibe, B. 2017.
\newblock In defense of the triplet loss for person re-identification.
\newblock \emph{arXiv preprint arXiv:1703.07737}.

\bibitem[{Ho, Jain, and Abbeel(2020)}]{15-ho2020denoising}
Ho, J.; Jain, A.; and Abbeel, P. 2020.
\newblock Denoising diffusion probabilistic models.
\newblock \emph{Advances in neural information processing systems}, 33: 6840--6851.

\bibitem[{Holmquist and Wandt(2023)}]{46-holmquist2023diffpose}
Holmquist, K.; and Wandt, B. 2023.
\newblock Diffpose: Multi-hypothesis human pose estimation using diffusion models.
\newblock In \emph{Proceedings of the IEEE/CVF international conference on computer vision}, 15977--15987.

\bibitem[{Huang et~al.(2023)Huang, Wang, Jin, Yang, He, Feng, and Liu}]{32-huang2023condition}
Huang, X.; Wang, X.; Jin, Z.; Yang, B.; He, B.; Feng, B.; and Liu, W. 2023.
\newblock Condition-adaptive graph convolution learning for skeleton-based gait recognition.
\newblock \emph{IEEE Transactions on image processing}.

\bibitem[{Li and Zhao(2022)}]{31-li2022strong}
Li, N.; and Zhao, X. 2022.
\newblock A strong and robust skeleton-based gait recognition method with gait periodicity priors.
\newblock \emph{IEEE Transactions on multimedia}, 25: 3046--3058.

\bibitem[{Li et~al.(2020)Li, Makihara, Xu, Yagi, Yu, and Ren}]{52-li2020end}
Li, X.; Makihara, Y.; Xu, C.; Yagi, Y.; Yu, S.; and Ren, M. 2020.
\newblock End-to-end model-based gait recognition.
\newblock In \emph{Proceedings of the asian conference on computer vision}.

\bibitem[{Liao et~al.(2020)Liao, Yu, An, and Huang}]{9-liao2020model}
Liao, R.; Yu, S.; An, W.; and Huang, Y. 2020.
\newblock A model-based gait recognition method with body pose and human prior knowledge.
\newblock \emph{Pattern recognition}, 98: 107069.

\bibitem[{Lin, Zhang, and Yu(2021)}]{51-lin2021gaitgl}
Lin, B.; Zhang, S.; and Yu, X. 2021.
\newblock Gait recognition via effective global-local feature representation and local temporal aggregation.
\newblock In \emph{Proc5eedings of the IEEE/CVF international conference on computer vision}, 14648--14656.

\bibitem[{Liu et~al.(2022)Liu, You, He, Bi, and Wang}]{62-liu2022symmetry}
Liu, X.; You, Z.; He, Y.; Bi, S.; and Wang, J. 2022.
\newblock Symmetry-Driven hyper feature GCN for skeleton-based gait recognition.
\newblock \emph{Pattern recognition}, 125: 108520.

\bibitem[{Loper et~al.(2023)Loper, Mahmood, Romero, Pons-Moll, and Black}]{10-loper2023smpl}
Loper, M.; Mahmood, N.; Romero, J.; Pons-Moll, G.; and Black, M.~J. 2023.
\newblock SMPL: A skinned multi-person linear model.
\newblock In \emph{Seminal graphics papers: pushing the boundaries, Volume 2}, 851--866.

\bibitem[{Min et~al.(2024)Min, Guo, Hao, and Dong}]{63-Min2024GaitMAPM}
Min, F.; Guo, S.; Hao, F.; and Dong, J. 2024.
\newblock GaitMA: Pose-guided Multi-modal Feature Fusion for Gait Recognition.
\newblock \emph{arXiv preprint arXiv:2407.14812}.

\bibitem[{Pan et~al.(2023)Pan, Chen, Xu, He, and He}]{33-pan2023toward}
Pan, H.; Chen, Y.; Xu, T.; He, Y.; and He, Z. 2023.
\newblock Toward complete-view and high-level pose-based gait recognition.
\newblock \emph{IEEE Transactions on information forensics and security}, 18: 2104--2118.

\bibitem[{Peng et~al.(2024)Peng, Ma, Zhang, and He}]{18-peng2024learning}
Peng, Y.; Ma, K.; Zhang, Y.; and He, Z. 2024.
\newblock Learning rich features for gait recognition by integrating skeletons and silhouettes.
\newblock \emph{Multimedia tools and applications}, 83(3): 7273--7294.

\bibitem[{Pinyoanuntapong et~al.(2023)Pinyoanuntapong, Ali, Wang, Lee, and Chen}]{53-pinyoanuntapong2023gaitmixer}
Pinyoanuntapong, E.; Ali, A.; Wang, P.; Lee, M.; and Chen, C. 2023.
\newblock Gaitmixer: skeleton-based gait representation learning via wide-spectrum multi-axial mixer.
\newblock In \emph{ICASSP 2023-2023 IEEE International conference on acoustics, speech and signal processing (ICASSP)}, 1--5. IEEE.

\bibitem[{Rombach et~al.(2022)Rombach, Blattmann, Lorenz, Esser, and Ommer}]{17-rombach2022high}
Rombach, R.; Blattmann, A.; Lorenz, D.; Esser, P.; and Ommer, B. 2022.
\newblock High-resolution image synthesis with latent diffusion models.
\newblock In \emph{Proceedings of the IEEE/CVF conference on computer vision and pattern recognition}, 10684--10695.

\bibitem[{Ronneberger, Fischer, and Brox(2015)}]{64-ronneberger2015u}
Ronneberger, O.; Fischer, P.; and Brox, T. 2015.
\newblock U-net: Convolutional networks for biomedical image segmentation.
\newblock In \emph{Medical image computing and computer-assisted intervention--MICCAI 2015: 18th international conference, Munich, Germany, October 5-9, 2015, proceedings, part III 18}, 234--241. Springer.

\bibitem[{Shen et~al.(2023)Shen, Fan, Wu, Wang, Huang, and Yu}]{41-shen2023lidargait}
Shen, C.; Fan, C.; Wu, W.; Wang, R.; Huang, G.~Q.; and Yu, S. 2023.
\newblock Lidargait: Benchmarking 3d gait recognition with point clouds.
\newblock In \emph{Proceedings of the IEEE/CVF conference on computer vision and pattern recognition}, 1054--1063.

\bibitem[{Song, Meng, and Ermon(2020)}]{16-song2020denoising}
Song, J.; Meng, C.; and Ermon, S. 2020.
\newblock Denoising diffusion implicit models.
\newblock \emph{arXiv preprint arXiv:2010.02502}.

\bibitem[{Takemura et~al.(2018)Takemura, Makihara, Muramatsu, Echigo, and Yagi}]{27-takemura2018multi}
Takemura, N.; Makihara, Y.; Muramatsu, D.; Echigo, T.; and Yagi, Y. 2018.
\newblock Multi-view large population gait dataset and its performance evaluation for cross-view gait recognition.
\newblock \emph{IPSJ Transactions on computer vision and applications}, 10: 1--14.

\bibitem[{Teepe et~al.(2022)Teepe, Gilg, Herzog, H{\"o}rmann, and Rigoll}]{7-teepe2022towards}
Teepe, T.; Gilg, J.; Herzog, F.; H{\"o}rmann, S.; and Rigoll, G. 2022.
\newblock Towards a deeper understanding of skeleton-based gait recognition.
\newblock In \emph{Proceedings of the IEEE/CVF conference on computer vision and pattern recognition}, 1569--1577.

\bibitem[{Teepe et~al.(2021)Teepe, Khan, Gilg, Herzog, H{\"o}rmann, and Rigoll}]{6-teepe2021gaitgraph}
Teepe, T.; Khan, A.; Gilg, J.; Herzog, F.; H{\"o}rmann, S.; and Rigoll, G. 2021.
\newblock Gaitgraph: Graph convolutional network for skeleton-based gait recognition.
\newblock In \emph{2021 IEEE international conference on image processing (ICIP)}, 2314--2318. IEEE.

\bibitem[{Vaswani et~al.(2017)Vaswani, Shazeer, Parmar, Uszkoreit, Jones, Gomez, Kaiser, and Polosukhin}]{37-vaswani2017attention}
Vaswani, A.; Shazeer, N.; Parmar, N.; Uszkoreit, J.; Jones, L.; Gomez, A.~N.; Kaiser, {\L}.; and Polosukhin, I. 2017.
\newblock Attention is all you need.
\newblock \emph{Advances in neural information processing systems}, 30.

\bibitem[{Wang, Chen, and Liu(2022)}]{57-wang2022frame}
Wang, L.; Chen, J.; and Liu, Y. 2022.
\newblock Frame-level refinement networks for skeleton-based gait recognition.
\newblock \emph{Computer cision and image understanding}, 222: 103500.

\bibitem[{Wang et~al.(2003)Wang, Tan, Ning, and Hu}]{2-wang2003silhouette}
Wang, L.; Tan, T.; Ning, H.; and Hu, W. 2003.
\newblock Silhouette analysis-based gait recognition for human identification.
\newblock \emph{IEEE Transactions on pattern analysis and machine intelligence}, 25(12): 1505--1518.

\bibitem[{Wang et~al.(2023{\natexlab{a}})Wang, Guo, Lin, Yang, Zhu, Li, Zhang, and Yu}]{44-wang2023dygait}
Wang, M.; Guo, X.; Lin, B.; Yang, T.; Zhu, Z.; Li, L.; Zhang, S.; and Yu, X. 2023{\natexlab{a}}.
\newblock DyGait: Exploiting dynamic representations for high-performance gait recognition.
\newblock In \emph{Proceedings of the IEEE/CVF international conference on computer vision}, 13424--13433.

\bibitem[{Wang et~al.(2023{\natexlab{b}})Wang, Hou, Zhang, Liu, Cao, and Huang}]{40-wang2023gaitparsing}
Wang, Z.; Hou, S.; Zhang, M.; Liu, X.; Cao, C.; and Huang, Y. 2023{\natexlab{b}}.
\newblock GaitParsing: Human semantic parsing for gait recognition.
\newblock \emph{IEEE Transactions on multimedia}.

\bibitem[{Wang et~al.(2024)Wang, Hou, Zhang, Liu, Cao, Huang, Li, and Xu}]{61-wang2024qagait}
Wang, Z.; Hou, S.; Zhang, M.; Liu, X.; Cao, C.; Huang, Y.; Li, P.; and Xu, S. 2024.
\newblock QAGait: Revisit Gait Recognition from a Quality Perspective.
\newblock In \emph{Proceedings of the AAAI conference on artificial intelligence}, volume~38, 5785--5793.

\bibitem[{Wu et~al.(2016)Wu, Huang, Wang, Wang, and Tan}]{1-wu2016comprehensive}
Wu, Z.; Huang, Y.; Wang, L.; Wang, X.; and Tan, T. 2016.
\newblock A comprehensive study on cross-view gait based human identification with deep cnns.
\newblock \emph{IEEE Transactions on pattern analysis and machine intelligence}, 39(2): 209--226.

\bibitem[{Yam, Nixon, and Carter(2004)}]{56-yam2004automated}
Yam, C.; Nixon, M.~S.; and Carter, J.~N. 2004.
\newblock Automated person recognition by walking and running via model-based approaches.
\newblock \emph{Pattern recognition}, 37(5): 1057--1072.

\bibitem[{Yu, Tan, and Tan(2006)}]{26-yu2006framework}
Yu, S.; Tan, D.; and Tan, T. 2006.
\newblock A framework for evaluating the effect of view angle, clothing and carrying condition on gait recognition.
\newblock In \emph{18th International conference on pattern recognition (ICPR'06)}, volume~4, 441--444. IEEE.

\bibitem[{Zhang et~al.(2023)Zhang, Chen, Han, and Liu}]{28-zhang2023spatial}
Zhang, C.; Chen, X.-P.; Han, G.-Q.; and Liu, X.-J. 2023.
\newblock Spatial transformer network on skeleton-based gait recognition.
\newblock \emph{Expert systems}, 40(6): e13244.

\bibitem[{Zheng et~al.(2023{\natexlab{a}})Zheng, Wu, Liu, Meng, and Zheng}]{24-zheng2023diffuvolume}
Zheng, D.; Wu, X.-M.; Liu, Z.; Meng, J.; and Zheng, W.-s. 2023{\natexlab{a}}.
\newblock Diffuvolume: Diffusion model for volume based stereo matching.
\newblock \emph{arXiv preprint arXiv:2308.15989}.

\bibitem[{Zheng et~al.(2022{\natexlab{a}})Zheng, Liu, Gu, Sun, Gan, Zhang, Liu, and Yan}]{45-zheng2022gait}
Zheng, J.; Liu, X.; Gu, X.; Sun, Y.; Gan, C.; Zhang, J.; Liu, W.; and Yan, C. 2022{\natexlab{a}}.
\newblock Gait recognition in the wild with multi-hop temporal switch.
\newblock In \emph{Proceedings of the 30th ACM international conference on multimedia}, 6136--6145.

\bibitem[{Zheng et~al.(2022{\natexlab{b}})Zheng, Liu, Liu, He, Yan, and Mei}]{21-zheng2022gait}
Zheng, J.; Liu, X.; Liu, W.; He, L.; Yan, C.; and Mei, T. 2022{\natexlab{b}}.
\newblock Gait recognition in the wild with dense 3d representations and a benchmark.
\newblock In \emph{Proceedings of the IEEE/CVF conference on computer vision and pattern recognition}, 20228--20237.

\bibitem[{Zheng et~al.(2023{\natexlab{b}})Zheng, Liu, Wang, Wang, Yan, and Liu}]{12-zheng2023parsing}
Zheng, J.; Liu, X.; Wang, S.; Wang, L.; Yan, C.; and Liu, W. 2023{\natexlab{b}}.
\newblock Parsing is all you need for accurate gait recognition in the wild.
\newblock In \emph{Proceedings of the 31st ACM international conference on multimedia}, 116--124.

\bibitem[{Zhu et~al.(2021)Zhu, Guo, Yang, Huang, Deng, Huang, Du, Lu, and Zhou}]{28-zhu2021gait}
Zhu, Z.; Guo, X.; Yang, T.; Huang, J.; Deng, J.; Huang, G.; Du, D.; Lu, J.; and Zhou, J. 2021.
\newblock Gait recognition in the wild: A benchmark.
\newblock In \emph{Proceedings of the IEEE/CVF international conference on computer vision}, 14789--14799.

\bibitem[{Zou et~al.(2024)Zou, Fan, Xiong, Shen, Yu, and Tang}]{60-zou2024cross}
Zou, S.; Fan, C.; Xiong, J.; Shen, C.; Yu, S.; and Tang, J. 2024.
\newblock Cross-Covariate Gait Recognition: A Benchmark.
\newblock In \emph{Proceedings of the AAAI conference on artificial intelligence}, volume~38, 7855--7863.

\end{thebibliography}

\title{Supplementary Material for ZipGait: Bridging Skeleton and Silhouette with Diffusion Model for Advancing Gait Recognition
}
\author{
    Written by AAAI Press Staff\textsuperscript{\rm 1}\thanks{With help from the AAAI Publications Committee.}\\
    AAAI Style Contributions by Pater Patel Schneider,
    Sunil Issar,\\
    J. Scott Penberthy,
    George Ferguson,
    Hans Guesgen,
    Francisco Cruz\equalcontrib,
    Marc Pujol-Gonzalez\equalcontrib
}
\setlist[itemize]{leftmargin=*}
\def\blue#1{\textcolor{blue}{#1}}

\renewcommand{\thefigure}{\Alph{figure}}
\renewcommand{\thetable}{\Alph{table}}
\renewcommand{\thealgorithm}{\Alph{algorithm}}

\setcounter{figure}{0}  
\setcounter{table}{0}   


\twocolumn[
  \maketitleB
]

\begin{figure*}[!ht]
    \setlength{\abovecaptionskip}{2pt}
    \setlength{\belowcaptionskip}{2pt}
    \begin{center}
        \includegraphics[width=1.0\linewidth]{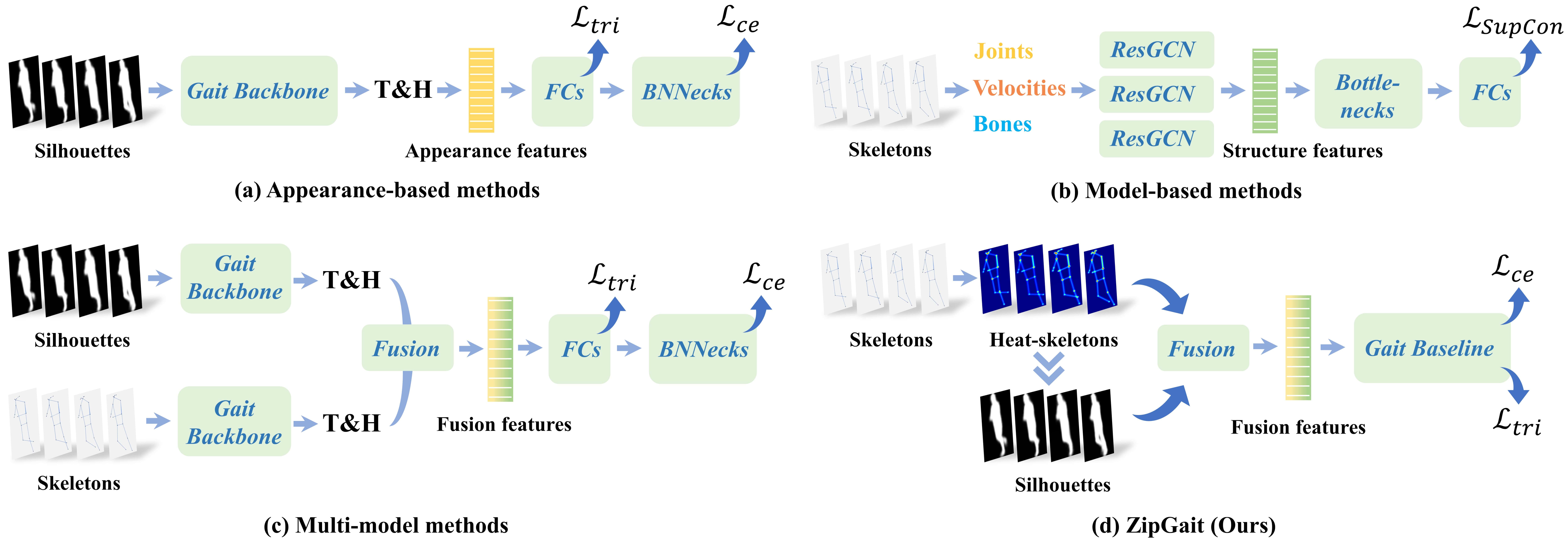}
    \end{center}
    \caption{\small Brief frameworks of related gait recognition methods, and our proposed ZipGait model. 
    By providing an overall description of the differences in the procedures of each method.
    \textbf{T\&H} represents the horizontal mapping and temporal aggregation.}
    \label{Pipeline-comparision}
\end{figure*}  

In appendix, we provide background on diffusion models, DiffGait training and inference algorithms, comparison with other related gait recognition methods, analysis of the impact on generalization and more visualization details in Section A, Section B, Section C, Section D and Section E.

\subsection{A. Background on Diffusion Models}
Diffusion models typically consist of two basic processes: 1) a forward process that gradually adds Gaussian noise to sample data, and 2) a reverse process that learns to invert the forward diffusion. 

Our work extends the framework of Denoising Diffusion Probabilistic Models (DDPMs), a class of deep generative models that approximate the distribution of natural images using the terminal state of a Markov chain that originates from a standard Gaussian distribution. DDPMs are trained to reverse a diffusion process that gradually adds Gaussian noise to the training data $x_{0}$ over $T$ steps until it becomes pure noise at $x_{T}$. 

To be specific, the forward process may be formalized as sampling from a conditional distribution $\mathrm{q}$, denoted as $\mathrm{q}\left(\mathbf{x}_t \mid \mathbf{x}_{t-1}\right)$, which mainly involves adding noise to the data without depending on any parameterized distribution,
\begin{eqnarray}
    \mathrm{q}\left(\mathbf{x}_{t} \mid \mathbf{x}_{t-1}\right)=\mathcal{N}\left(\mathbf{x}_{t} ; \sqrt{1-\beta_{t}} \mathbf{x}_{t-1}, \beta_{t} \mathbf{I}\right),
\end{eqnarray}
where $\mathbf{I}$ is the identity matrix and the rate at which the original data is diffused into noise is controlled by a variance scheduling given by $\beta_{1}$, . . . , $\beta_{T}$. Importantly, the forward process and its schedule may be reparameterized as:
\begin{eqnarray}
    \alpha_t=\prod_{t=1}^T\left(1-\beta_t\right),
\end{eqnarray}
\begin{eqnarray}
    \mathrm{q}\left(\mathbf{x}_t \mid \mathbf{x}_0\right)=\mathcal{N}\left(\mathbf{x}_t ; \sqrt{\alpha_t} \mathbf{x}_{t-1},\left(1-\alpha_t\right) \mathbf{I}\right).
\end{eqnarray}

In contrast, the reverse process reconstructs the data distribution from noise, represented as $\mathrm{p}\left(\mathbf{x}_{t-1} \mid \mathbf{x}_{t}\right)$.
This reverse conditional distribution is arbitrarily complex, but we might approximate it via a denoising deep neural network,
\begin{eqnarray}
    \mathrm{p}\left(\mathbf{x}_{t-1} \mid \mathbf{x}_t\right) \approx \mathcal{N}\left(\mathbf{x}_{t-1} ; \mu_\theta\left(\mathbf{x}_t, t\right), \Sigma_t\right).
\end{eqnarray}
where $\mu_\theta\left(\mathbf{x}_t, t\right)$ is a learned deep neural network that has as input both the noisy data $x_{t}$ and its step t (usually position encoded); $\Sigma_t$ depends on the variance schedule but is not otherwise learned.

The conditional diffusion models, which add an extra term in the denoising process, provide the opportunity for the cross-modal generation tasks:
\begin{eqnarray}
    \mathrm{p}\left(\mathbf{x}_{t-1} \mid \mathbf{x}_t, c\right) \approx \mathcal{N}\left(\mathbf{x}_{t-1} ; \mu_\theta\left(\mathbf{x}_t, c, t\right), \Sigma_t\right),
\end{eqnarray}
where the conditioning inputs $\mathrm{c}$ added to $\epsilon_\theta$ may be derived from the original modality. 

\subsection{B. DiffGait Training and Inference Algorithms}

\noindent \textbf{Training.} During the training phase, we perform the diffusion process that corrupts ground truth silhouette $DG_0$ to noisy silhouette $DG_t$, then $G_{ske}$ is integrated with $DG_t$ to $HGV_t$. 
We learn the reverse denoising process by continuously using $\mathcal{D}$ to recover $HGV_t$ into $DG_0$, thus optimizing \textit{DiffGait} to gradually remove the uncertainty of $DG_t$ and redefine the source distribution $DG_0$ to $P_t$, which can be formulated as:
\begin{equation}
    P_t=\mathcal{D}(HGV_t).
\end{equation}
We employ MSE to supervise the model training:
\begin{equation}
    \mathcal{L}_{dg}=\left\| DG_0-P_t \right\|_{2}^{2}.
\end{equation}
Algorithm~\ref{alg:training} provides the overall training procedure.

\noindent \textbf{Inference.}
Algorithm~\ref{alg:inference} summarizes the detailed inference procedure of \textit{DiffGait}, which can be seen as iteratively generating more complete silhouettes. 
We have redesigned the sampling method based on DDIM to generate effective silhouettes at each sampling step, observing that later-generated features are more complete yet tend to lack some details.
Specifically, for each sampling step, $\mathcal{D}$ takes the initial $HGV_T$ or the $HGV_t$ from the previous step as input and outputs the estimated silhouette for the current step.
Then, our new sampling method is used to update the silhouette for the next step.

\begin{algorithm}
    \footnotesize
    \caption{DiffGait Training}
    \label{alg:training}
    \begin{algorithmic}[1]
        \REQUIRE Heat-skeleton : $\mathcal{I}_t$, GT\_Silhouette : $DG_0$
        \REPEAT
        \STATE $G_{ske} = \mathcal{E}(\mathcal{I}_t)$
        \STATE $t \sim \text{Uniform}(\{1, \dots, T\})$
        \STATE $\epsilon \sim \mathcal{N}(0, 1)$
        \STATE $DG_t = \sqrt{\alpha_t}DG_0 + \sqrt{1-\alpha_t}\epsilon$
        \STATE $HGV_t=G_{ske}\odot(GM(DG_{t})+TE(t)) + G_{ske}$
        \STATE Take gradient descent step on \\
        $\Delta \theta \|\mathcal{D}(HGV_t) - DG_0\|^2$
        \UNTIL{converged}
    \end{algorithmic}
    
\end{algorithm}

\begin{algorithm}
    \footnotesize
    \caption{DiffGait Inference}
    \label{alg:inference}
    \begin{algorithmic}[1]
        \REQUIRE Heat-skeleton : $\mathcal{I}_t$, steps : $T$
        \ENSURE Predicted\_Silhouette : $P_i$
        \STATE $DG_T \sim \mathcal{N}(0, 1)$
        \STATE $G_{ske} = \mathcal{E}(\mathcal{I}_t)$
        \STATE $times = \text{Reversed}(\text{Linspace}(-1, T, steps))$
        \STATE $time\_pairs = \text{List}(\text{Zip}(times[:-1], times[1:]))$
        \STATE $HGV_T=G_{ske}\odot(GM(DG_{T})+TE(T)) + G_{ske}$
        \FOR{$(t_{\text{now}}, t_{\text{next}})$ in time\_pairs}
            \STATE $P_i = \mathcal{D}(HGV_T, t_\text{now})$
            \STATE $HGV_t = \textit{DiffGait}(HGV_T, P_ii, t_{\text{now}}, t_{\text{next}})$
        \ENDFOR
        \RETURN $P_t$
    \end{algorithmic}
\end{algorithm}

\subsection{C. Comparison with Other Related Gait Recognition Methods}
In the aforementioned article, we briefly delineate the differences between our ZipGait approach and other common gait recognition methods in terms of input modalities and feature utilization. To concretely highlight the distinctions of our architecture, we have illustrated four pipelines in Fig \ref{Pipeline-comparision}, each representing the overall workflow of our method compared to three other approaches in gait recognition.

As shown in Fig~\ref{Pipeline-comparision}(a), appearance-based methods are currently the mainstream algorithms in the field of gait recognition. This paradigm, followed by most studies, typically exhibits the best performance under general conditions. However, these methods inevitably suffer from the quality of silhouettes; in environments with complex backgrounds or significant occlusions, performance is limited, as low-quality silhouettes do not enable the models to learn robust gait representations, which is a very critical factor.

Fig~\ref{Pipeline-comparision}(b) illustrates model-based approaches, which primarily extract gait features from human body models. The skeleton is the only model that can be used as a standalone input for gait recognition, often referred to as skeleton-based methods. SMPL models and point clouds are usually employed as supplementary modalities to provide additional information. The main limitation of model-based methods is the accuracy of the models, especially three-dimensional ones, which currently cannot effectively estimate the absolute posture of the human body in three-dimensional space, making gait recognition using these models unreliable.

The basic workflow of multimodal methods is depicted in Fig~\ref{Pipeline-comparision}(c), which extracts features from different modalities through two branches and then merges them for feature classification. This approach does offer a novel strategy for obtaining more effective gait representations, but it requires inputs from multiple modalities and the processing of features from these modalities, which is particularly demanding in terms of computational and storage resources. Although this can achieve high performance, further efforts are needed to make the structure more lightweight.

Fig~\ref{Pipeline-comparision}(d) presents our ZipGait method, which borrows the advantages of the three aforementioned approaches while addressing some of their shortcomings. Our method solely uses the skeleton as an input but establishes a relationship with the silhouette through our DiffGait model. In terms of feature fusion, we have refined the dual-branch structure of current multimodal methods by merging features at a lower dimension, effectively simplifying the model structure. Overall, ZipGait employs a highly streamlined structure, using only one modality as input, yet achieving the effects of multi-modality.

It must be noted that our proposed method still lags behind appearance-based and multimodal methods in performance. This is primarily due to the predicted silhouettes not fully restoring all external details, tending instead to reconstruct a more averaged distribution.

\subsection{D. Analysis of the Impact on Generalization}
\begin{table}[!t]
    \setlength{\abovecaptionskip}{2pt}
    \setlength{\belowcaptionskip}{2pt}
    \caption{\small Ablation study on the effect of \textit{DiffGait} reconstructed silhouettes on generalization capability. Heat. denotes Heat-skeleton; Silh. donates Silhouette;
    The best performances are in \textbf{blod}.}
    \label{ablation: diffgait-branch-cross}
    \footnotesize
    \centering
    \begin{tabular}{cc|cc|cc}
        \toprule
        \multirow{2}{*}{Heat.} & \multirow{2}{*}{Silh.} & \multicolumn{2}{c|}{Trained on CASIA-B} & \multicolumn{2}{l}{Trained on Gait3D} \\
        \cline{3-6}& & CASIA-B & Gait3D & CASIA-B & Gait3D \\
        \hline 
        $\checkmark$ & \ding{55} & 77.2 & 10.2 & 27.4 & 33.6 \\
        \ding{55} & $\checkmark$ & 67.1 & 9.8 & 34.5 & 24.7 \\
        $\checkmark$ & $\checkmark$ & \textbf{83.4} & \textbf{12.8} & \textbf{44.2} & \textbf{39.5} \\       
        \bottomrule
    \end{tabular}
\end{table}

Our proposed ZipGait outperforms other methods in cross-domain evaluations. To ascertain whether the enhanced generalization ability is due to the use of DiffGait in reconstructing silhouettes, we specifically designed experiments to separately assess the intra-domain and cross-domain evaluation effects of Heat-skeleton and predicted Silhouette. The experiments demonstrate that incorporating DiffGait is a key factor in improving the model's generalization capability.

\begin{figure}[!ht]
    \setlength{\abovecaptionskip}{2pt}
    \setlength{\belowcaptionskip}{2pt}
    \begin{center}
        \includegraphics[width=1\linewidth]{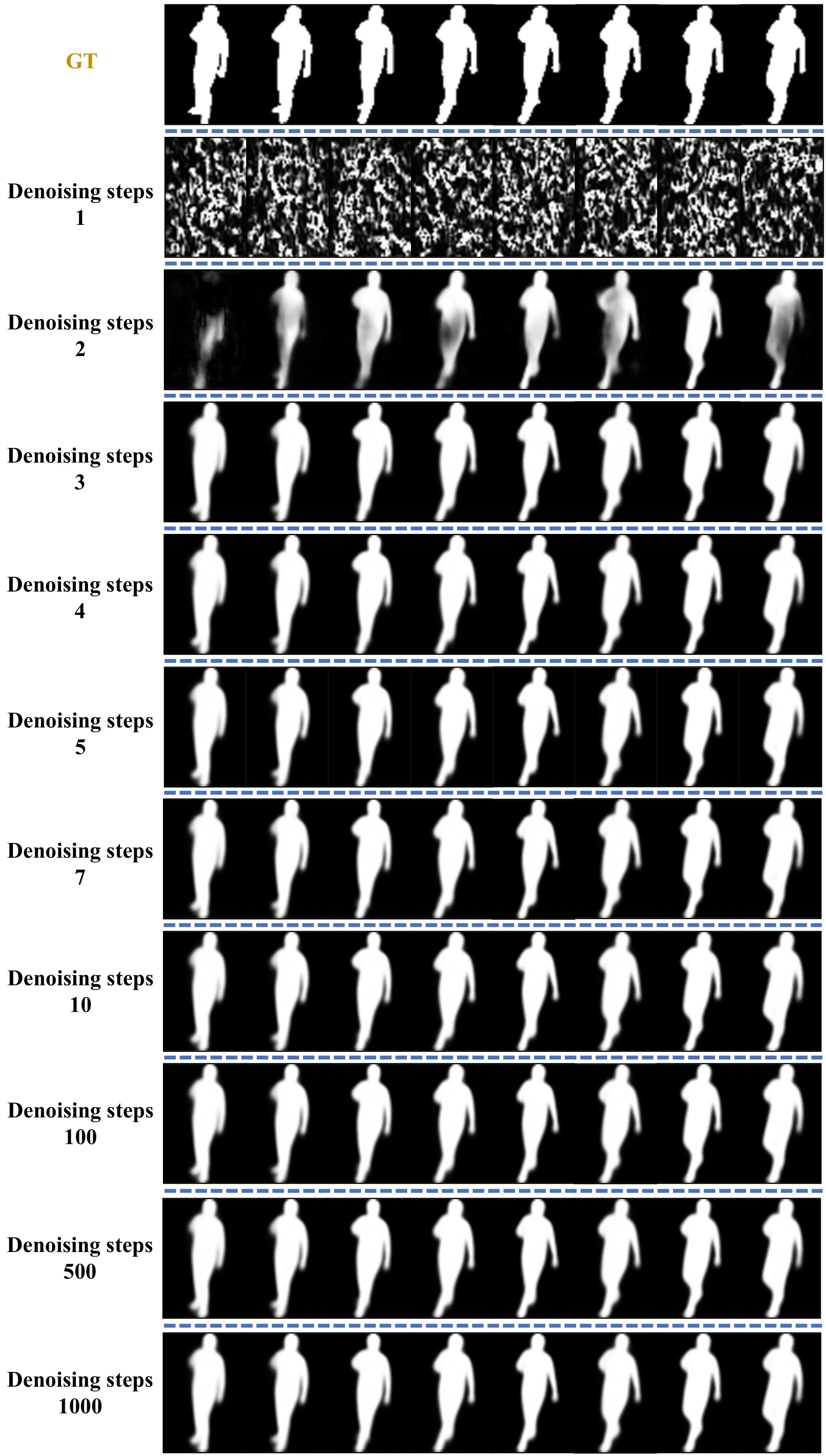}
    \end{center}
    \caption{\small Visualization of the reconstructed results under different denoising steps of DiffGait on Gait3D.}
    \label{visualization_denoising_steps}
\end{figure}  

\begin{figure}[!t]
    \setlength{\abovecaptionskip}{2pt}
    \setlength{\belowcaptionskip}{2pt}
    \begin{center}
        \includegraphics[width=1\linewidth]{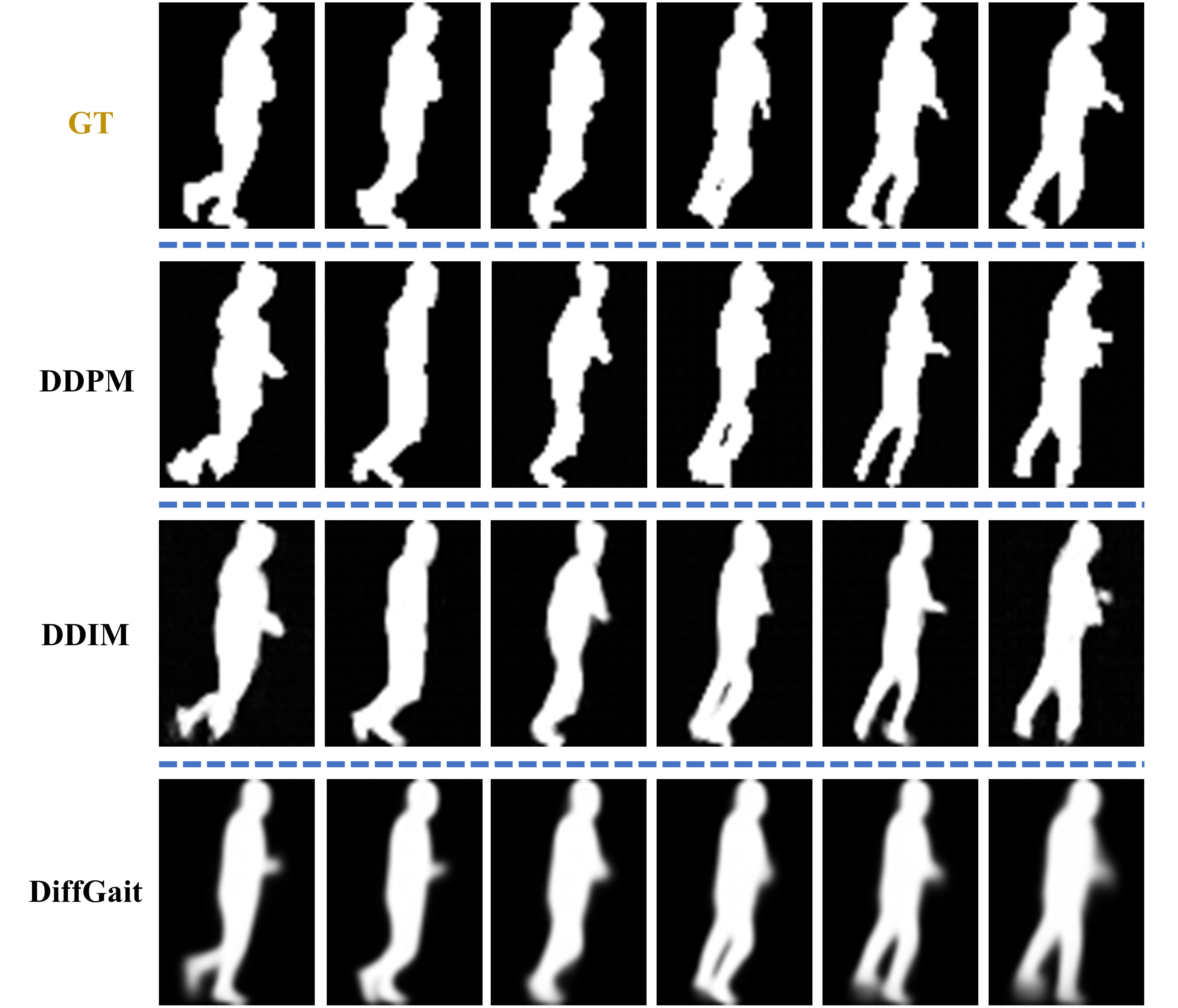}
    \end{center}
    \caption{\small Visualization of the silhouette reconstruction results of our DiffGait, DDIM, DDPM in one gait sequence.}
    \label{visualization_model_denoising}
\end{figure}  

\begin{figure}[!t]
    \setlength{\abovecaptionskip}{2pt}
    \setlength{\belowcaptionskip}{2pt}
    \begin{center}
        \includegraphics[width=1\linewidth]{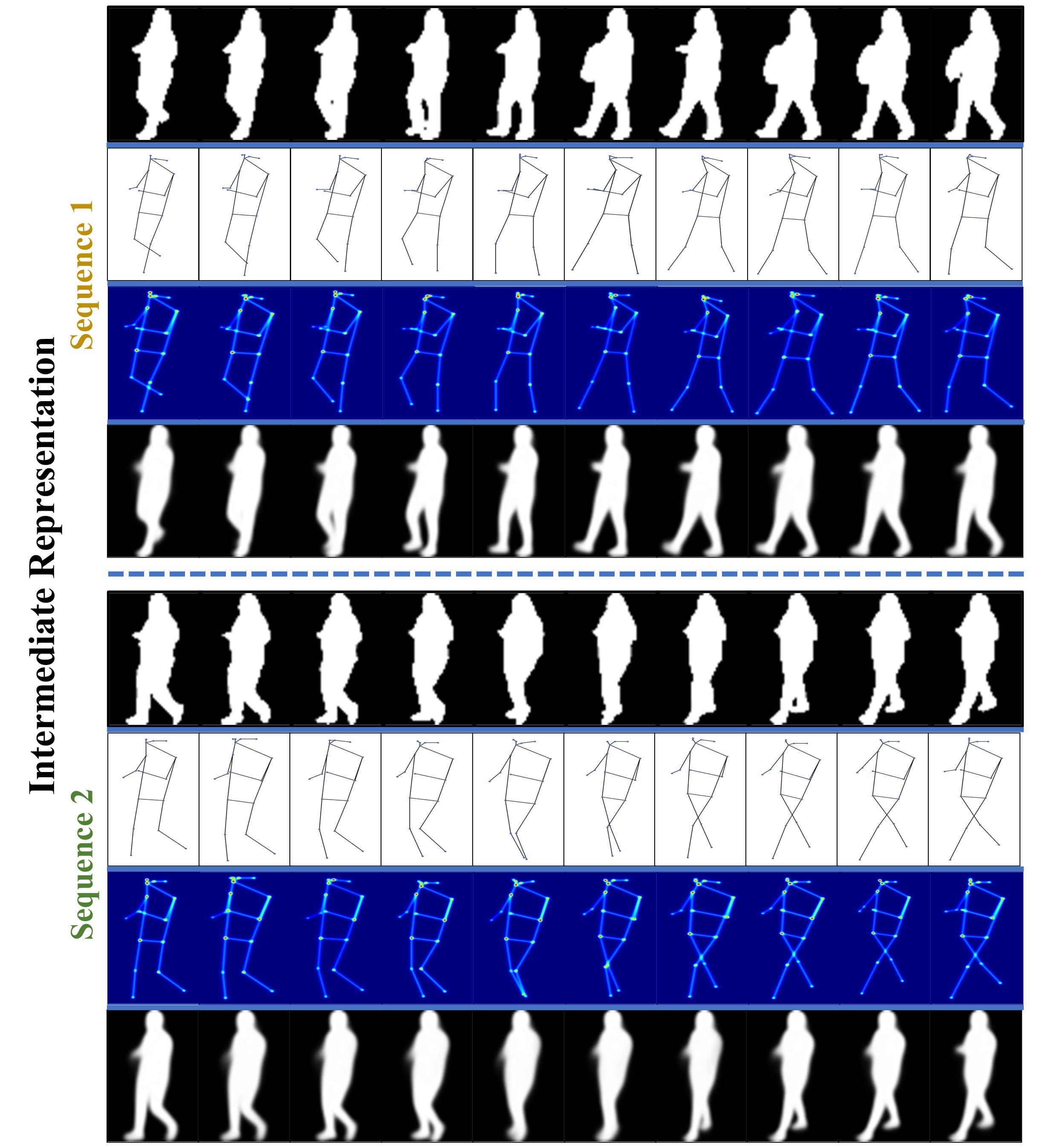}
    \end{center}
    \caption{\small Visual results of our DiffGait on different intermediate representations in the whole process.}
    \label{intermediate_representation}
\end{figure}  

\subsection{E. More Visualization Details.}
\begin{figure*}[!ht]
    \setlength{\abovecaptionskip}{2pt}
    \setlength{\belowcaptionskip}{2pt}
    \begin{center}
        \includegraphics[width=1\linewidth]{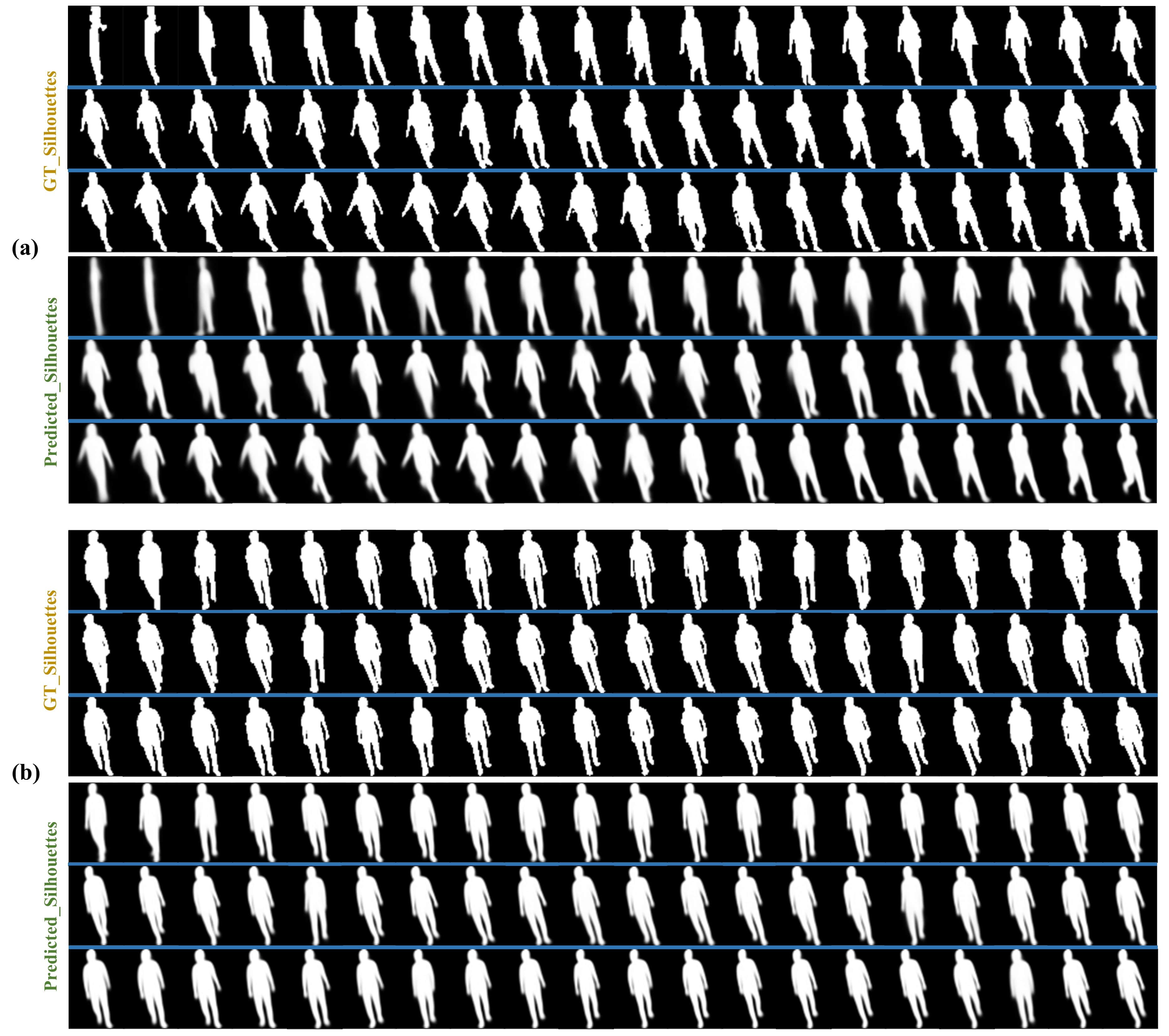}
    \end{center}
    \caption{\small Visual results of our DiffGait on different long sequences.}
    \label{visualization_full_sequence}
\end{figure*}

\begin{figure*}[!ht]
    \setlength{\abovecaptionskip}{2pt}
    \setlength{\belowcaptionskip}{2pt}
    \begin{center}
        \includegraphics[width=1\linewidth]{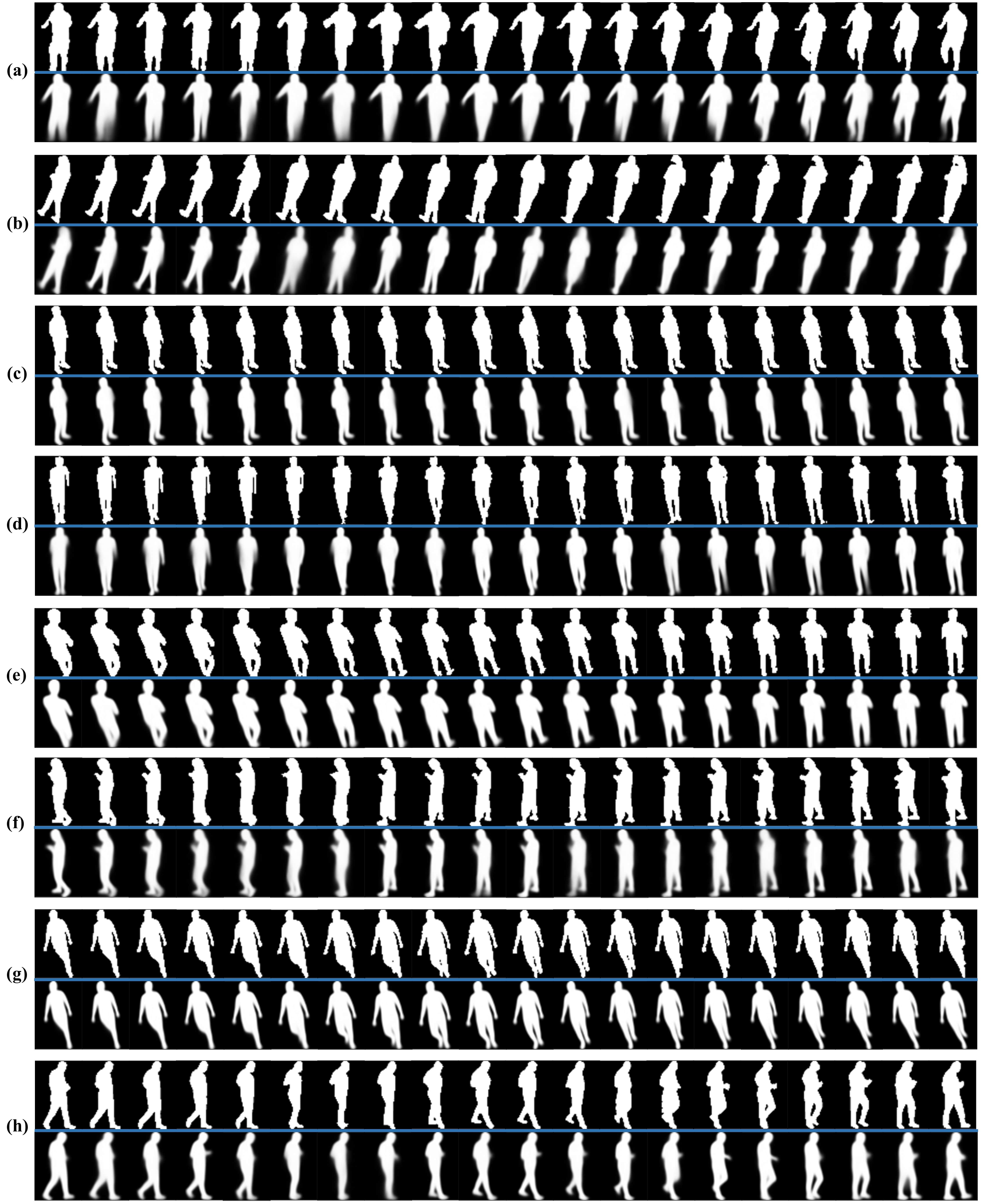}
    \end{center}
    \caption{\small Visual results of our DiffGait on Gait3D. Normal scenes with relatively complete shape information.}
    \label{visualization_normal}
\end{figure*}

\begin{figure*}[!ht]
    \setlength{\abovecaptionskip}{2pt}
    \setlength{\belowcaptionskip}{2pt}
    \begin{center}
        \includegraphics[width=1\linewidth]{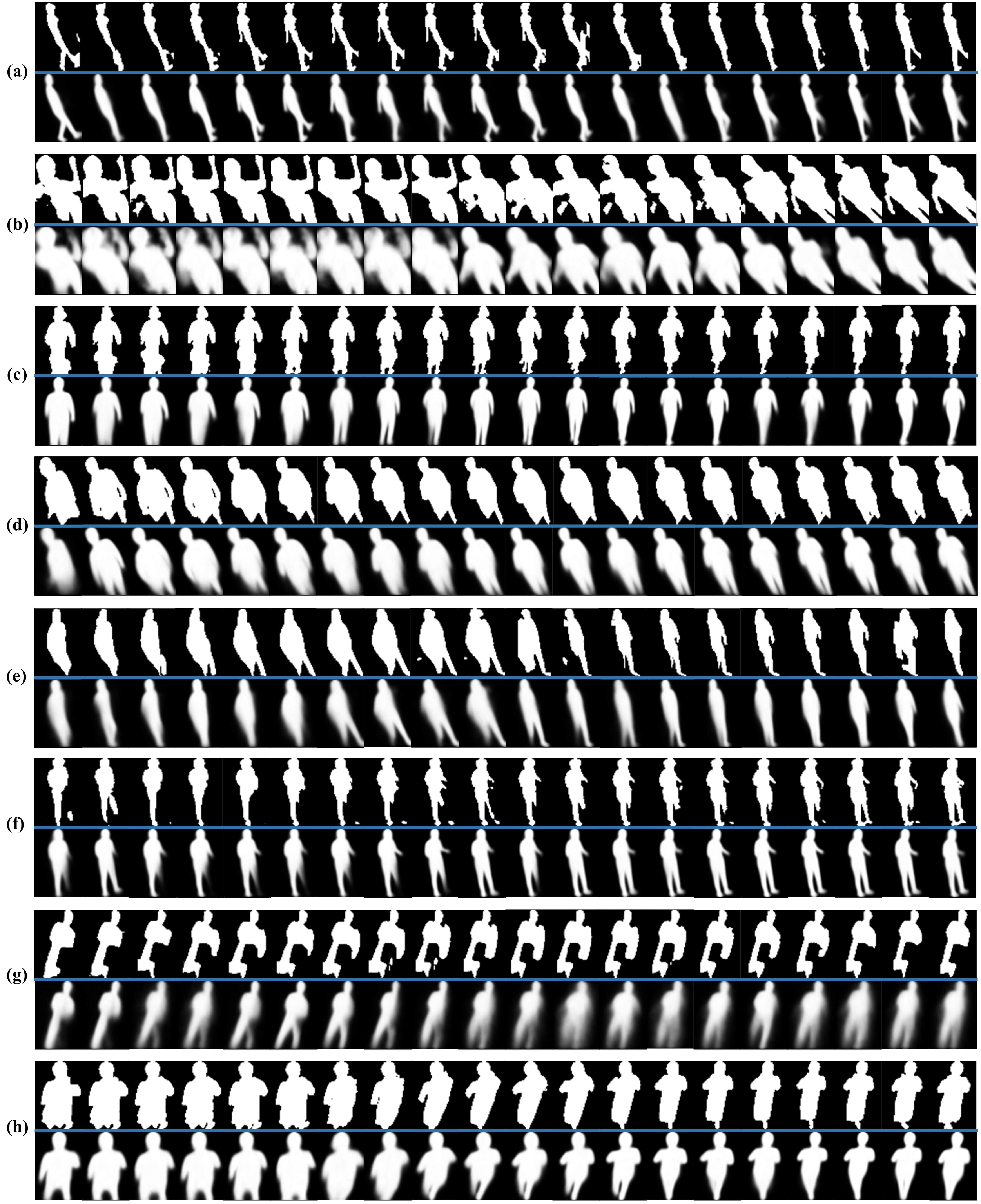}
    \end{center}
    \caption{\small Visual results of our DiffGait on Gait3D. Challenging scenes such as arbitrary views or occlusions are involved.}
    \label{visualization_difficult}
\end{figure*}

In this section, we provide more comprehensive visualizations to demonstrate the effectiveness of our proposed DiffGait model.
Fig.~\ref{visualization_denoising_steps} displays the denoising results at different denoising steps, showing that visually effective silhouettes can be achieved when the denoising step is set to three or more,.
Fig.~\ref{visualization_model_denoising} compares the performance with other diffusion models, revealing that our model does not significantly differ in denoising quality from the DDIM and DDPM models, and that our DiffGait is far more efficient than these methods. 
Fig.~\ref{intermediate_representation} visualizes the intermediate modalities used in ZipGait: skeleton, Heat-skeleton, and reconstructed silhouette, along with the ground truth silhouette.
Fig.~\ref{visualization_full_sequence} compares the predicted silhouettes of two long sequences (each with 60 frames) derived from our DiffGait.
Fig.~\ref{visualization_normal} presents visualizations under normal conditions, where there is enough shape information to reconstruct high-quality contours. 
Fig.~\ref{visualization_difficult} shows visualizations under challenging conditions, where the silhouettes reconstructed by DiffGait provide more effective features, as the robustness displayed by the skeleton allows it to address the impacts of occlusions.

These sufficient visualizations evaluate the quality of the DiffGait reconstructed silhouettes from different aspects and can fully demonstrate the effectiveness of our method.

\end{document}